\documentclass {article} 
\usepackage{arxiv}
\usepackage{booktabs} 
\usepackage{algorithm}%
\usepackage{algpseudocode}%
\usepackage{hyperref}
\usepackage{amsmath}
\usepackage{xcolor}
\usepackage{xspace}
\usepackage{soul}
\newcommand{\ignore}[1]{}
\usepackage[textsize=tiny]{todonotes}
\usepackage{graphicx}
\usepackage{boldline}
\usepackage[export]{adjustbox}
\usepackage[caption=false]{subfig}
\usepackage[font=small]{caption}
\usepackage{adjustbox}
\usepackage{enumerate}
\usepackage{amsfonts}
\usepackage{colortbl}
\usepackage{float}
\usepackage{rotating}
\usepackage{setspace}
\usepackage[official]{eurosym} 
\usepackage[flushleft]{threeparttable}
\usepackage{enumitem}
\usepackage{array}
\title{An Evolutionary Deep Learning Method for  Short-term Wind Speed Prediction: A Case Study of the Lillgrund Offshore Wind Farm}

\author{
 Mehdi Neshat \\
  Optimization and Logistics Group\\
  School of Computer Science\\
  The University of Adelaide\\
   Australia \\
  \texttt{mehdi.neshat@adelaide.edu.au} \\
   \And
   Meysam Majidi Nezhad\\
Department of Astronautics\\
Electrical and Energy Engineering (DIAEE)\\
	Sapienza University of Rome\\
	 Italy\\
	 \texttt{meysam.majidinezhad@uniroma1.it} \\
   \And
Ehsan Abbasnejad \\
  The Australian Institute for Machine Learning\\
   The University of Adelaide\\
   Australia \\
  \texttt{ehsan.abbasnejad@adelaide.edu.au} \\
  \And
  Lina Bertling Tjernberg \\
  School of Electrical Engineering and Computer Science\\
  KTH Royal Institute of Technology Stockholm\\
  Sweden\\
    \texttt{Linab@kth.se} \\
  \And
  Davide Astiaso Garcia \\
  Department of Planning, Design\\
  and Technology of Architecture\\
  Sapienza University of Rome\\
  Italy\\ 
    \texttt{davide.astiasogarcia@uniroma1.it} \\
  \And
  Bradley Alexander \\
  Optimization and Logistics Group\\
  School of Computer Science\\
  The University of Adelaide\\
   Australia \\
  \texttt{bradley.alexander@adelaide.edu.au} \\
  \And
  Markus Wagner \\
  Optimization and Logistics Group\\
  School of Computer Science\\
  The University of Adelaide\\
   Australia \\
  \texttt{markus.wagner@adelaide.edu.au} 
  }

\begin{document}

\maketitle
\doublespacing
\begin{abstract}

Accurate short-term wind speed forecasting is essential for large-scale integration of wind power generation. However, the seasonal and stochastic characteristics of wind speed make forecasting a challenging task. This study uses a new hybrid evolutionary approach that uses a popular evolutionary search algorithm, CMA-ES, to tune the hyper-parameters of two  long-term-short-term (LSTM) ANN models for wind prediction. 

The proposed hybrid approach is trained on data gathered from an offshore wind turbine installed in a Swedish wind farm located in the Baltic Sea. Two forecasting horizons including ten-minutes ahead (absolute short term) and one-hour ahead (short term) are considered in our experiments. Our experimental results indicate that the new approach is superior to five other applied machine learning models, i.e., polynomial neural network (PNN), feed-forward neural network (FNN), nonlinear autoregressive neural network (NAR) and adaptive neuro-fuzzy inference system (ANFIS), as measured by  five performance criteria.

\end{abstract}


\sloppy

\section{Introduction}\label{sec:Introduction}

Concerning trajectories for global warming and environmental pollution have motivated intensive efforts to replace fossil fuels\cite{nezhad2019wind}.
One of the most important clean energy sources is wind. Wind energy has key advantages in terms of technological maturity, cost, and life-cycle greenhouse gas emissions\cite{nezhad2018nearshore}.
However, wind is a variable resource, so
accurate wind power forecasts are crucial in reducing the incidence of costly curtailments, and in protecting system integrity and worker-safety\cite{lerner2009importance}.
However, obtaining an accurate local wind speed prediction can be difficult. This is because the wind speed characteristics are stochastic, intermittent and non-stationary which can 
defeat simple models\cite{khare2016solar}.

In this paper, we propose a hybrid evolutionary deep forecasting model combining a recurrent deep learning model (LSTM network)\cite{schmidhuber1997long}, coupled with the CMA-ES algorithm\cite{hansen2004evaluating}, (called CMAES-LSTM) for predicting the short-term wind speed with high accuracy. 

  As there is no straightforward theory governing the design of an LSTM network for a given problem\cite{hu2018nonlinear}, we tune model structure and hyper-parameters using a combination of grid search and CMA-ES. 
    We demonstrate the performance of the proposed hybrid (CMAES-LSTM) 
  model using a real case study using data collected from the Lillgrund offshore wind-farm to predict wind speeds ten-minutes ahead and one hour ahead. 
   The proposed method is compared with the FNN model, ANFIS model, PNN model, NAR model and a static LSTM model. Statistical analyses show that the proposed adaptive method exhibits better performance than these current (static) models.

The remainder this article is structured as follows. The next section briefly surveys related work in the field of predictive wind models. Section~\ref{sec:Methodology} presents current methodologies, theories and our proposed hybrid evolutionary deep learning (CMAES-LSTM) model. Section~\ref{sec:performance} exhibits the performance indices applied for evaluating the introduced models. After this, Section~\ref{sec:case_study} gives a brief description of the offshore wind farm applied in this paper. Section~\ref{sec:Experiments} describes and  analyses our experimental results. Lastly, we provide a summary and outline future work in Section~\ref{sec:conclusion}.

\section{Related Work}
Today, wind turbine generators (WTGs) are installed in onshore, nearshore and offshore areas worldwide\cite{luo2018wind,vachaparambil2014optimal,us2013biological,natarajan2014sustainability}.
Sweden is one of the leading countries harnessing offshore wind power due to its geographical location, access to shallow seas and strong North winds on the Baltic Sea. 
Wind energy is variable wind resources that are influenced by several factors including: the location of the turbines (on-, near-, offshore); turbine height; seasons, meso-scale and diurnal variations; and climate change. All of which can  affect the stable operation of the power grid~\cite{xia2018optimal}. Studies have shown how these factors can impact on the the reliability of wind energy\cite{akccay2017short}.

The short-term predictive models of wind speed based on historical data have been developed using autoregressive moving average models~\cite{erdem2011arma}, autoregressive integrated moving average models~\cite{yu2018data}.
Related work by Gani et. al.\cite{gani2016combined} combined non-linear models with support vector machines. 
Wind forecast methodologies using ANNs include Elman neural networks \cite{yu2017improved}, polynomial neural networks (PNN)\cite{zjavka2015wind}, feed-forward neural networks (FNN)\cite{masrur2016short} and long short-term memory Network (LSTM)\cite{hu2018nonlinear}, hybrid artificial neural network\cite{wang2016wind}. 

Recently deep-learned ANNs solutions for wind forecasting have proven popular. Hu et al.~\cite{hu2016transfer} used transfer learning for short term prediction.
Wang et al.~\cite{wang2016deep} introduced a new wind speed forecasting approach using a deep belief network (DBN) based on the deterministic and probabilistic variables. 
Liu used recurrent deep learning models based on Long Short Term Memory (LSTM) network for forecasting wind speed in different time-scales \cite{liu2018wind,liu2018smart}. Chen et al. ~\cite{chen2018wind} recommended an ensemble of six different LSTM networks configurations for wind speed forecasting with ten-min and one-hour interval.
More broadly  hybrid nonlinear forecasting models have been explored for the prediction of wind energy generation, solar energy forecasting and energy market  forecasting~\cite{zhang2013short,khodayar2018spatio,lin2018multi,vladislavleva2013predicting,moreno2018wind,belaid2016prediction}.

The work in this paper differs from the previous work in wind-prediction in the use of global heuristic search methods to optimise both network structure and tune hyper-parameters. Our approach customises known methodologies from neuro-evolution\cite{stanley2019designing} to improve the performance of predictive wind models. 
\section{Methodology} \label{sec:Methodology}
In this section, we introduce the proposed methodologies and related concepts, including LSTM network details, CMA-ES and the combined  LSTM network and the CMA-ES algorithm.

\subsection{Long short-term memory network (LSTM)}
An LSTM is a type of recurrent neural network (RNN) which has the capacity to model time series data with different long-term and short-term dependencies~\cite{schmidhuber1997long}. The core of the LSTM network is the memory cell, which regards the hidden layers place of the of traditional neurons LSTM is equipped with three gates (input, output and forget gates), therefore it is able to add or remove information to the cell state. For calculating the estimated outputs and updating the state of the cell, the following equations can be used: 
\begin{equation}\label{eq-LSTM1}
i_t=\sigma (W_{ix}x_t+W_{im}m_{t-1}+W_{ic}c_{t-1}+b_i)
\end{equation}
\begin{equation}\label{eq-LSTM2}
i_t=\sigma (W_{fx}x_t+W_{fm}m_{t-1}+W_{fc}c_{t-1}+b_f)
\end{equation}
\begin{equation}\label{eq-LSTM3}
c_t=f \odot c_{t-1}+i_t\odot g(W_{cx}x_t+W_{cm}m_{t-1}+b_c)
\end{equation}
\begin{equation}\label{eq-LSTM4}
o_t=\sigma (W_{ox}x_t+W_{om}m_{t-1}+W_{oc}c_{t}+b_o)
\end{equation}
\begin{equation}\label{eq-LSTM5}
m_t=o_t \odot h(c_t)
\end{equation}
\begin{equation}\label{eq-LSTM6}
y_t=W_{ym} m_t+b_y
\end{equation}
where $x_t$ is the input and $y_t$ is the output;  $i_t$, $o_t$ and $f_t$ indicate the input gate, output gate and forget gate. The activation
vectors of each cell is shown by $c_t$, while $m_t$ denotes the activation vectors for any memory block. $\sigma$, $g$ and $h$ express the activation function of the gate, input and output (the logistic sigmoid and $tanh$ function are assigned). Lastly, $\odot$ (Hadamard product) indicates the element-wise multiplication between two vectors. Furthermore, $b_i$, $b_f$,$b_c$, $b_o$ are the corresponding bias vectors. $W_{ox}$,$W_{om}$, $W_{oc}$, $W_{ix}$, $W_{im}$, $W_{ic}$, $W_{fx}$, $W_{fm}$, $W_{fc}$, $W_{cx}$, $W_{cm}$, and $W_{ym}$ are the corresponding weight coefficients.

\subsection{Covariance Matrix Adaptation Evolution Strategy (CMA-ES)}

The CMA-ES \cite{hansen2004evaluating} search process for an  $n$-dimensional problem works by adapting an $n\times n$ covariance-matrix $C$ which defines the shape and orientation of a Gaussian distribution in the search space and a vector $x$ that describes the location of the centre of the distribution. Search is conducted by sampling this distribution for a population of $\mu$ individual solutions. These solutions are then evaluated and the relative performance of these solutions is used to update both $C$ and $x$. This process of sampling and adaptation continues until search converges or a fixed number of iterations has expired.

CMA-ES relies on three principal operations, which are selection, mutation, and recombination. Recombination and mutation are employed for exploration of the search space and creating genetic variations, whereas the operator of selection is for exploiting and converging to an optimal solution. The mutation operator plays a significant role in CMA-ES, which utilizes a multivariate Gaussian distribution. For a thorough explanation of the different selection operators, we refer the interested reader to~\cite{beyer2002evolution}.
 
CMA-ES can explore and exploit search spaces due to its self-adaptive mechanism for setting the vector of mutation step sizes ($\sigma$) instead of having just one global mutation step size. 
Self-adaptation can also improve convergence speed~\cite{hansen2004evaluating}. 
The covariance matrix is computed based on the differences in the mean values of two progressive generations. In which case, it expects that the current population includes sufficient information to estimate the correlations. After calculating the covariance matrix, the rotation matrix will derive from the covariance matrix with regard to expanding the distribution of the multivariate Gaussian in the estimated direction of the global optimum. This can be accomplished by conducting an eigen-decomposition of the covariance matrix to receive an orthogonal basis for the matrix~\cite{hansen2014principled}.
\subsection{Adaptive Tuning Process}
 One of the primary challenges in designing an ANN is setting appropriate values for the hyper-parameters such as the number of the hidden layers, number of neurons in each  layer, batch size, learning rate and type of the optimizer\cite{hu2018nonlinear}.  Tuning the hyper-parameters plays a significant role in improving the performance of the DNN with respect to problem domain. In the domain of wind forecasting Chen\cite{chen2018wind} has noted that the forecasting accuracy of LSTM networks influenced by structural parameters. There are three main techniques for tuning hyper-parameters. These include 1) manual trial and error which is costly and cannot be practised adequately, 2) systematic grid search, and 3) meta-heuristic search. In this paper, we compare the performance of both grid search and a meta-heuristic approach (CMAES-LSTM) in tuning LSTM networks for wind forecasting. 

 \begin{table} 
 \small
\caption{Summary of the predictive models tested in this paper. }
\centering
\label{table:details_model}
\scalebox{0.99}{
\begin{tabular}{|l|p{9cm}|}
\hline 
 
\textbf{Models} & \textbf{Descriptions}\\ \hline\hline
 LSTM \cite{hu2018nonlinear} + grid search& Long Short-term memory Network:
 \begin{itemize}
 \item \textbf{LSTM hyper-parameters}
 
 \begin {itemize}
 
 \item \textit{miniBatchSize}=512
 \item \textit{LearningRate}= $10^{-3}$
 \item \textit{numHiddenUnits1}   = 125;
\item \textit{numHiddenUnits2}   = 100;
\item \textit{Epochs} = 100
\item \textit{Optimizer}= 'adam'

\end{itemize}

 \end{itemize}
 \\ \hline
 ANFIS \cite{pousinho2011hybrid}& Adaptive neuro-fuzzy inference system:
 \begin{itemize}
 \item  OptMethod= Backpropagation
   \item \textbf{Training settings}

  \begin{itemize}
 
\item  \textit{Epochs}=100;
\item  \textit{ErrorGoal}=0;
\item  \textit{InitialStepSize}=0.01;
\item  \textit{StepSizeDecrease}=0.9;
\item  \textit{StepSizeIncrease}=1.1;
\end{itemize} 
 
 \item \textbf{FIS features}
  \begin{itemize}
 
\item  \textit{mf} number=5;
\item  \textit{mf} type='gaussmf';

\end{itemize} 
 
 \end{itemize}
    \\  \hline
 PNN \cite{zjavka2015wind}& Polynomial neural network:
 \begin{itemize}
 \item \textbf{PNN parameters}
  \begin {itemize}
  \item \textit{MaxNeurons}=20
 \item \textit{MaxLayers}= 5
 \item \textit{SelectionPressure}= 0.2;
\item \textit{TrainRatio}= 0.8;
\end{itemize}
 \end{itemize}
 \\ \hline
 FNN \cite{masrur2016short}& 
 Feed-forward neural network
 \begin{itemize}
 \item \textbf{FNN settings}
  \begin {itemize}
  \item \textit{hiddenSizes}= 100
 \item \textit{hiddenLayers}= 2
 \item \textit{trainFcn}= 'trainlm';
\end{itemize}
 \end{itemize}
 
 \\ \hline 
 NAR \cite{lydia2016linear}& 
 Nonlinear autoregressive neural network 
 (is similar to FNN settings)
 
 \\ \hline 
 CMAES-LSTM &  
 
 \begin{itemize}
 \item \textbf{CMAES-LSTM hyper-parameters (Best configuration)}
 
 \begin {itemize}
 
 \item \textit{miniBatchSize}=$655$
 \item \textit{LearningRate}=$10^{-3}$ 
 \item \textit{numHiddenUnits1}=$177$ ;
\item \textit{numHiddenUnits2}=$151$ ;
\item \textit{Epochs} = 100
\item \textit{Optimizer}= 'adam'
\item \textit{PopulationSize}=12
\item \textit{MaxEvaluation} =1000
\end{itemize}

 \end{itemize}
 \\ \hline
\end{tabular}
}

\end{table}

In the grid search method, we assign a fixed value for the optimizer type ('adam') \cite{kingma2014adam}, the number of LSTM hidden layers and also the number of neurons to the values in Table \ref{table:details_model}. The grid search process determines the batch size and learning rate can to within ranges of ($10^{-5} \le LR \le 10^{-1}$, and $8 \le BS \le 1024$). For search using CMA-ES we all of the listed hyper-parameters of the LSTM networks listed in the corresponding section of Table~\ref{table:details_model}.In order to avoid search just converging toward complex network designs that take too long to train we add penalty term for model training time to the fitness function $f$.
We frame the optimisation process as:

\begin{align}
\label{eq:fitness1}
\begin{split}
Argmin \to  f&=fitness(N_{h_1},N_{h_2},...,N_{h_D},N_{n_1h_1},N_{n_2h_2},...N_{n_Dh_D},L_R,B_S,Op) ,
\\
 Subject-to&:\\
 LN_h &\le N_h\le UN_h,\\
 LN_n &\le N_n\le UN_n,\\
10^{-5}&\le L_R \le 10^{-1},\\
8 &\le B_S \le 1024.
\end{split}
\end{align}
where $N_{h_i}, \{i=1,\ldots, D\}$ is the number of hidden layers for the $i-^{th}$ LSTM network and $N_{n_i,h_j}, \{j=1,\ldots, D_l\}$ is the number of neurons in the $i^{th}$ hidden  layer of this network. The lower and upper bounds of $N_h$ are shown by $LN_h$ and $UN_h$ , while $LN_n$ and  $UN_n$ are the lower and upper bounds of neuron number. The final fitness function is:
\begin{align}
\label{eq:fitness2}
\begin{split}
 f&=f_1+\omega f_2 \\
 f_1&=RMSE=\sqrt{\frac{1}{N}\sum_{i=1}^{N}(f_p(i)-f_o(i))^2}
 \end{split}
\end{align}
\begin{equation} \label{eq:fitness3}
\mathit{f_2}=
\begin {cases}
Tr_{runtime}-\rho, & if(Tr_{runtime}> \rho)\\
0, & \mathit{otherwise}
\end{cases}
\end{equation}
where RMSE is the root mean square error of the test samples; $\rho$ is the threshold of training runtime by 600(s). $\omega$ is the weight coefficient to penalise long training times. In this work we set $\omega$ to $10^{-3}$ because the range of RMSE is between $0.5$ and $1.5$. The value $N$ is the number of test data. Using the results of recent investigations of the lower and upper bounds the $N_h$ and $N_n$ which represent the desirable learning performance of LSTM networks \cite{nakisa2018long, rahman2018predicting, neshat2019adaptive}, we set the upper and lower bounds of $N_h$ and $N_n$ in the range $\{1,2\}$ and $\{30,230\}$ respectively. 

The optimisation process is initialised by providing CMA-ES with the search space bounds above and starting parameters of $\sigma=0.25$ and $\lambda=12$ (population size). Meanwhile the hyper-parameters of other forecasting models are initialised by the references. 

 \begin{figure*}[tbp]
  \includegraphics[width=\textwidth]{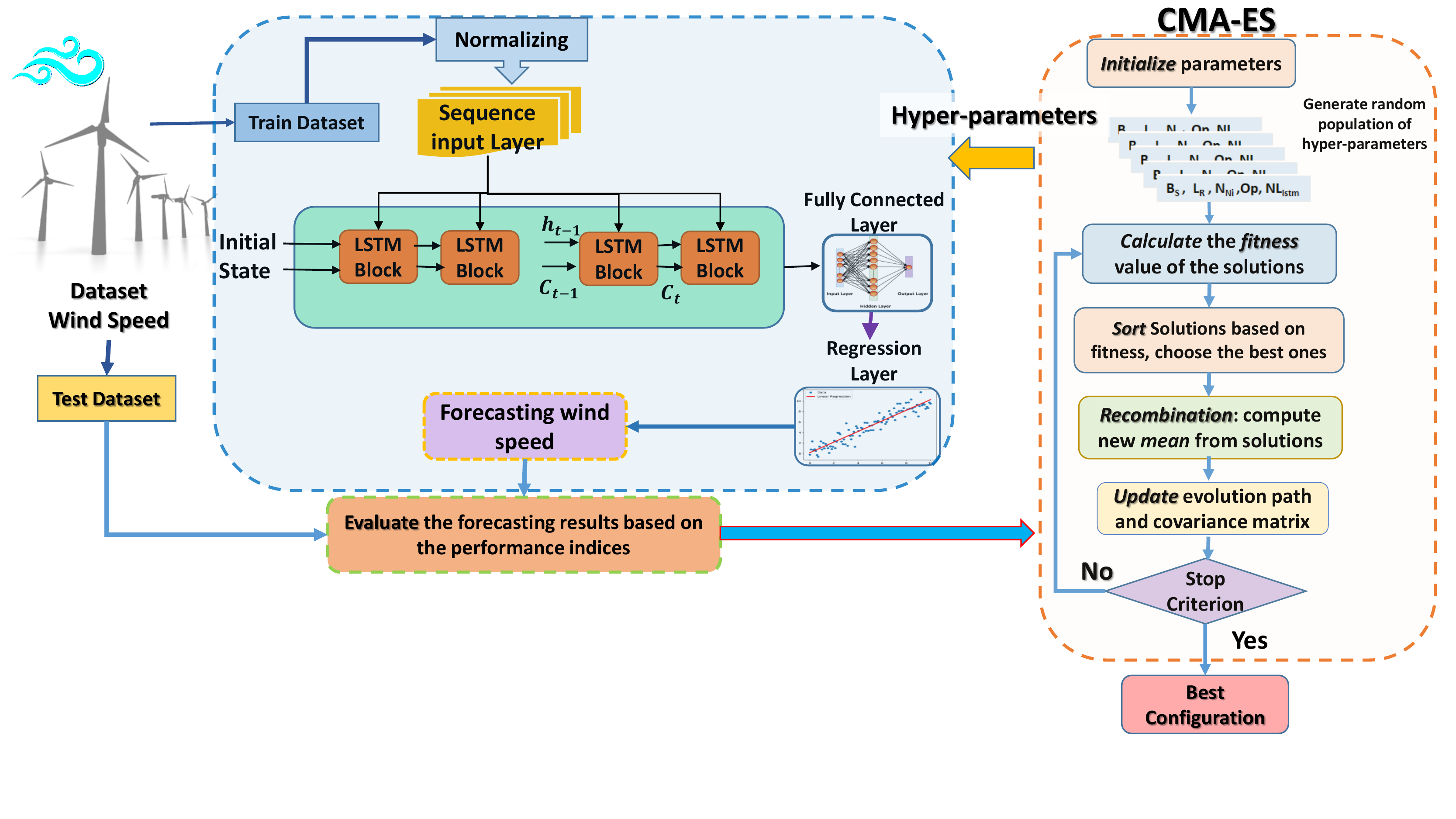}
   \caption{The forecasting framework of the proposed hybrid CMAES-LSTM model.  }
   \label{fig:cmaes_lstm}
  \end{figure*}
Figure~\ref{fig:cmaes_lstm} shows the overall optimization process of our hybrid model.
\section{Performance indices of forecasting models}
\label{sec:performance}
We use five broadly considered performance to assess the forecasting performance: the mean square error (MSE), the root mean square error (RMSE), mean absolute error (MAE), the mean absolute percentage error (MAPE) and the Pearson correlation coefficient (R) \cite{zhang2017compound}. 
The equations of MAE, RMSE, MAPE and R are represented as follows :
\begin{equation}\label{eq:MAE}
MAE=\frac{1}{N}\sum_{i=1}^{N}|f_p(i)-f_o(i)|
\end{equation}
\begin{equation}\label{eq:RMSE}
RMSE=\sqrt{\frac{1}{N}\sum_{i=1}^{N}(f_p(i)-f_o(i))^2}
\end{equation}
\begin{equation}\label{eq:MAPE}
MAPE=\frac{1}{N}\sum_{i=1}^{N}\frac{(f_p(i)-f_o(i))}{f_o(i)}\times100\%
\end{equation}
\begin{equation}\label{eq:R}
R=\frac{\frac{1}{N}\sum_{i=1}^{N}(f_p(i)-\overline{f}_p)(f_o(i)-\overline{f}_o)}
{\sqrt{\frac{1}{N}\sum_{i=1}^{N}(f_p(i)-\overline{f}_p)^2}\times\sqrt{\frac{1}{N}\sum_{i=1}^{N}(f_o(i)-\overline{f}_o)^2}}
\end{equation}
where $f_p(i)$ and $f_o(i)$ signify the predicted and observed wind speed values at the $i^{th}$ data point. The total number of observed data points in $N$. In addition, $\overline{f}_p$ and $\overline{f}_o$ are the average of the projected and observed consequences, respectively. For improving the performance of the predicted model, MSE, RMSE, MAE and MAPE should be minimised, while R needs to be maximized. 

 \begin{figure*}[tbp]
 \centering
  \includegraphics[width=0.9\textwidth]{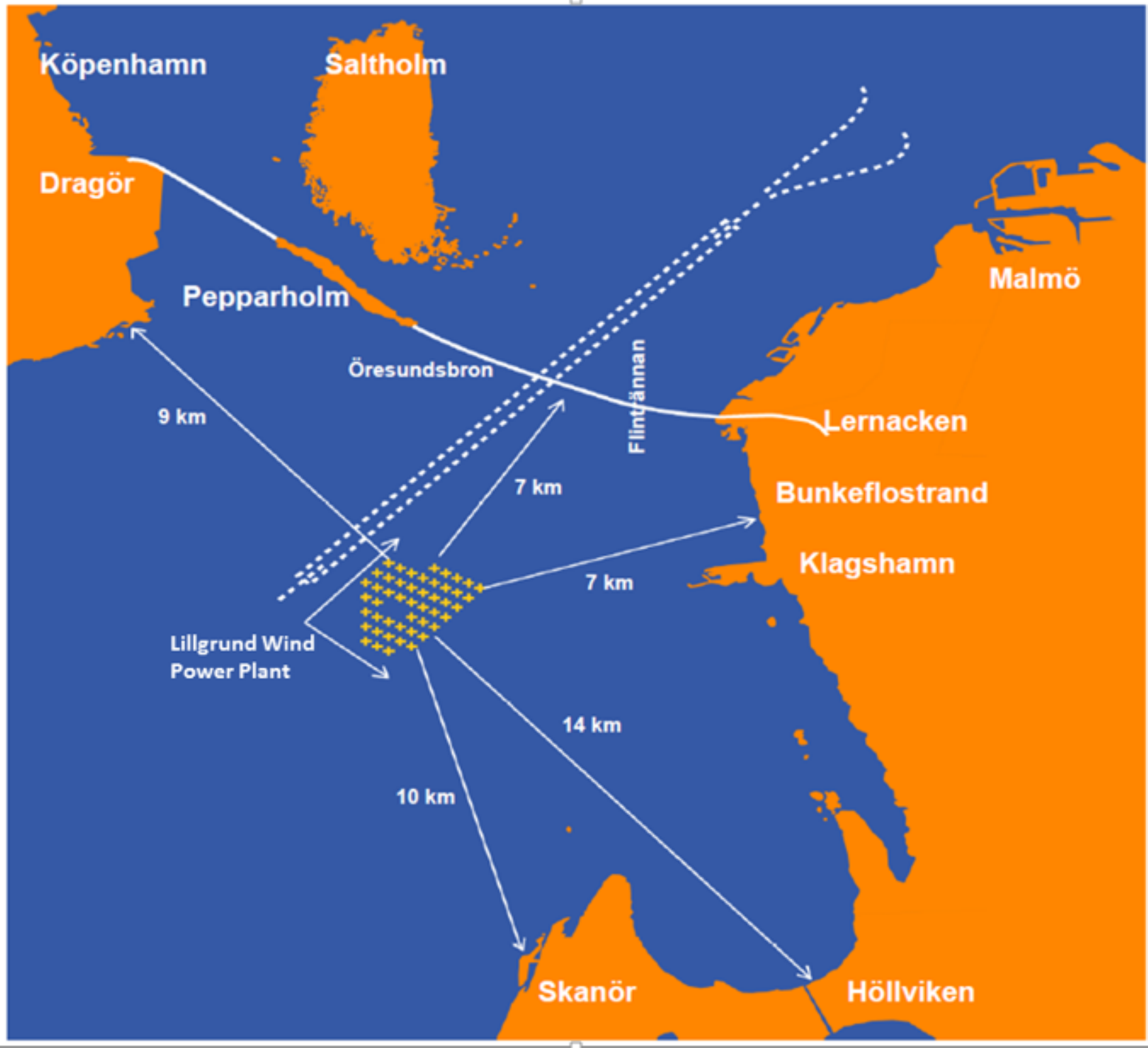}
   \caption{Location of the Lillgrund offshore wind power plant \cite{jeppsson2008technical}.}
   \label{fig:map}
 
  \end{figure*}
\section{Case Study}
\label{sec:case_study}
In this paper, we use the original wind speed data gathered from a large offshore wind farm called Lillgrund \cite{jeppsson2008technical}
, which is situated in a shallow area of Oresund, located 7 km off the coast of Sweden and 7 km south from the Oresund Bridge connecting Sweden and Denmark (see Figure \ref{fig:map}). The mean wind speed is around 8,5 m/s at hub height. This wind, together with the low water depth of 4 to 8 (m), makes the installation of wind turbines economically feasible. The Lillgrund offshore wind farm consists of 48 wind turbines, each rated at 2.3 (MW), resulting in a total wind power plant potential of 110 (MW)~\cite{goccmen2016estimation}. A SCADA collects wind power plant information at a 10-minute interval \cite{Dahlberg2009}. The wind power system also includes an offshore substation, an onshore substation and a 130 (kV) sea and land cable for connecting to shore.

 \begin{figure*}[tbp]
  \includegraphics[width=\textwidth]{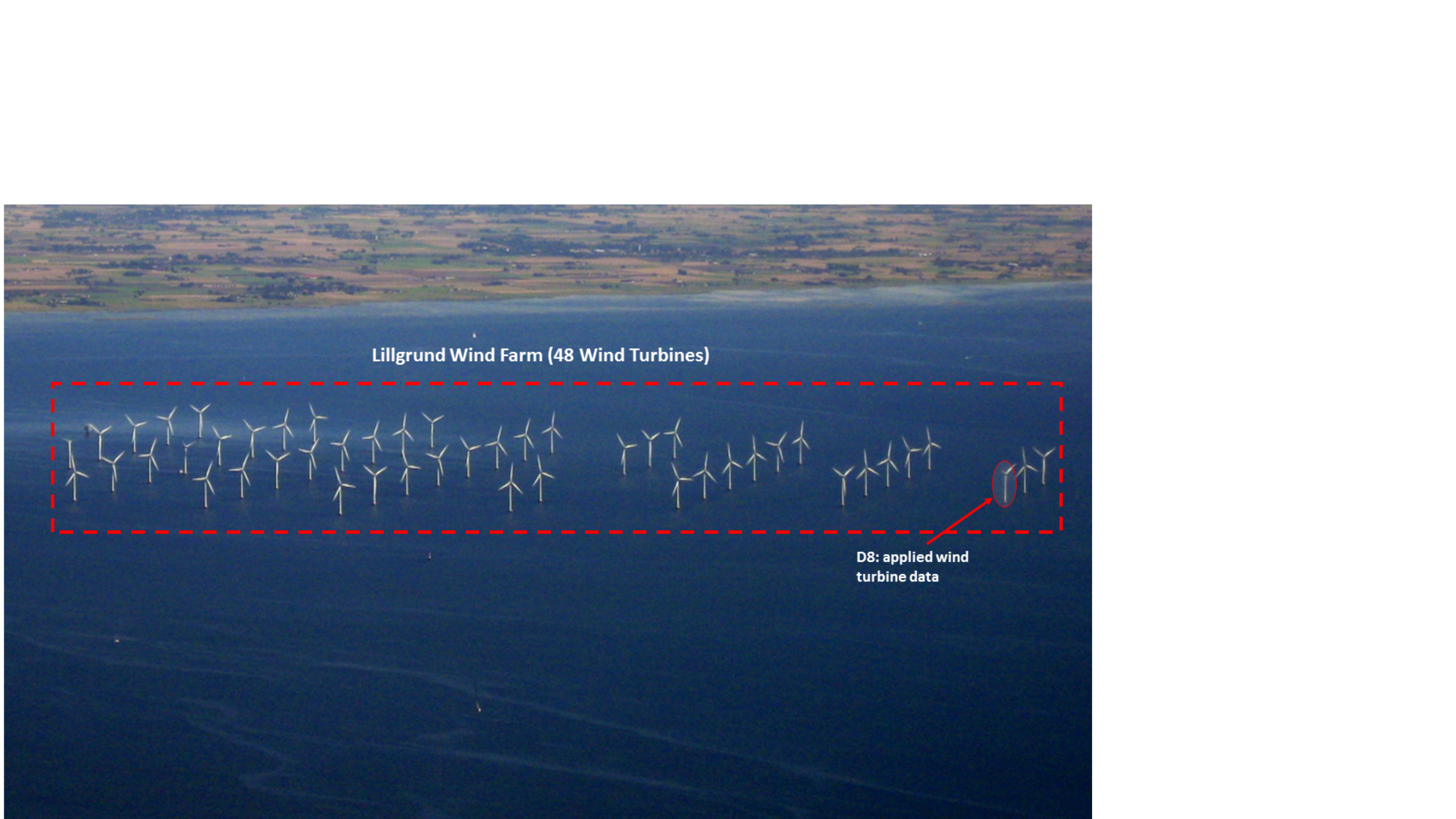}
   \caption{Lillgrund offshore wind farm in Baltic Sea and the wind turbine position is showed with red cycle which is applied for collecting the real wind speed data \cite{jeppsson2008technical}.}
   \label{fig:positionD3}
 
  \end{figure*}

\begin{figure}[t]
\centering
\includegraphics[clip,width=0.85\columnwidth]{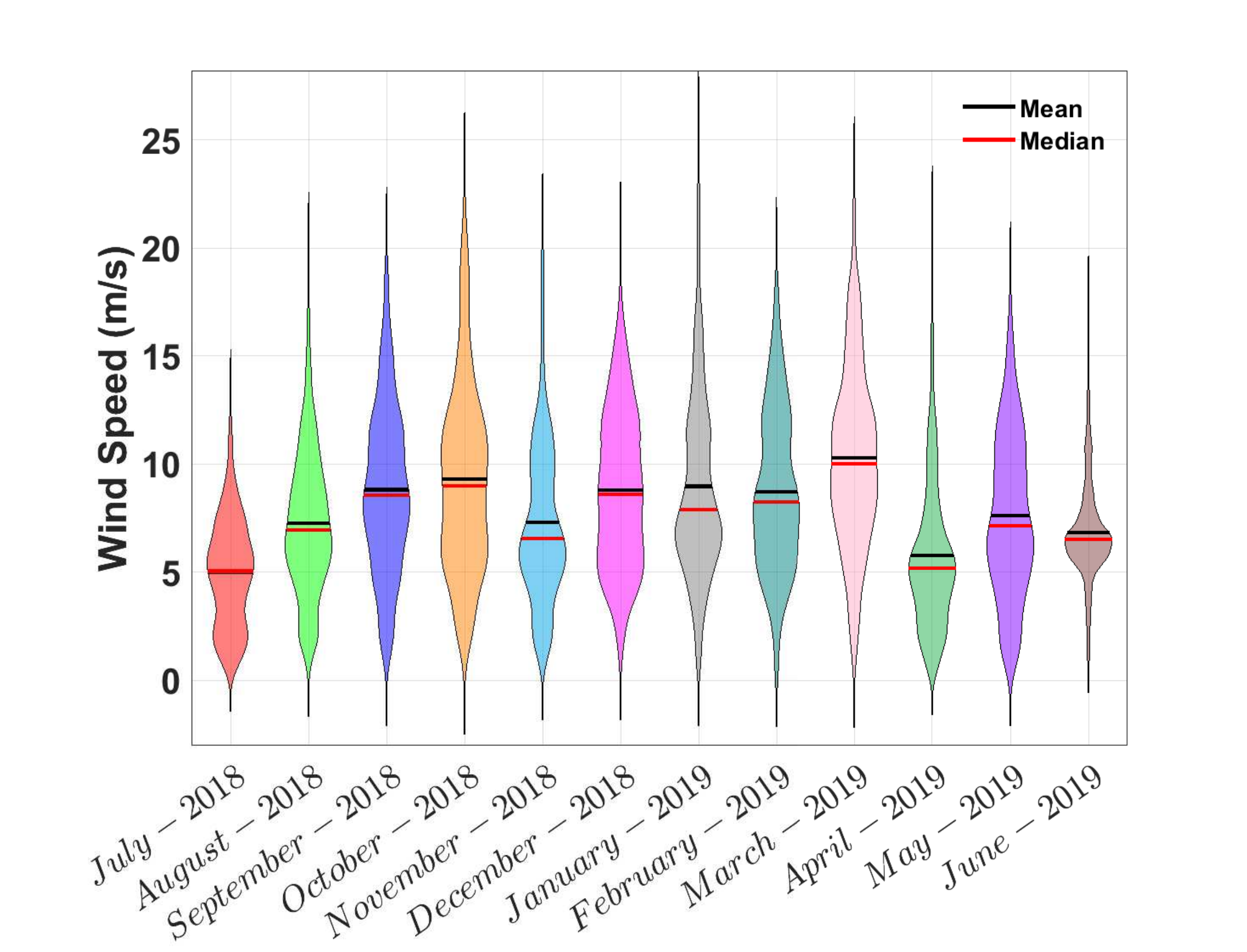}
\caption{the distribution and frequency of the wind speed data in Lillgrund Wind coastal site per 12 months. }%
\label{fig:boxplot_speed}
\end{figure}

 \begin{figure*}[tbp]
 \centering
  \includegraphics[clip,width=0.6\columnwidth]{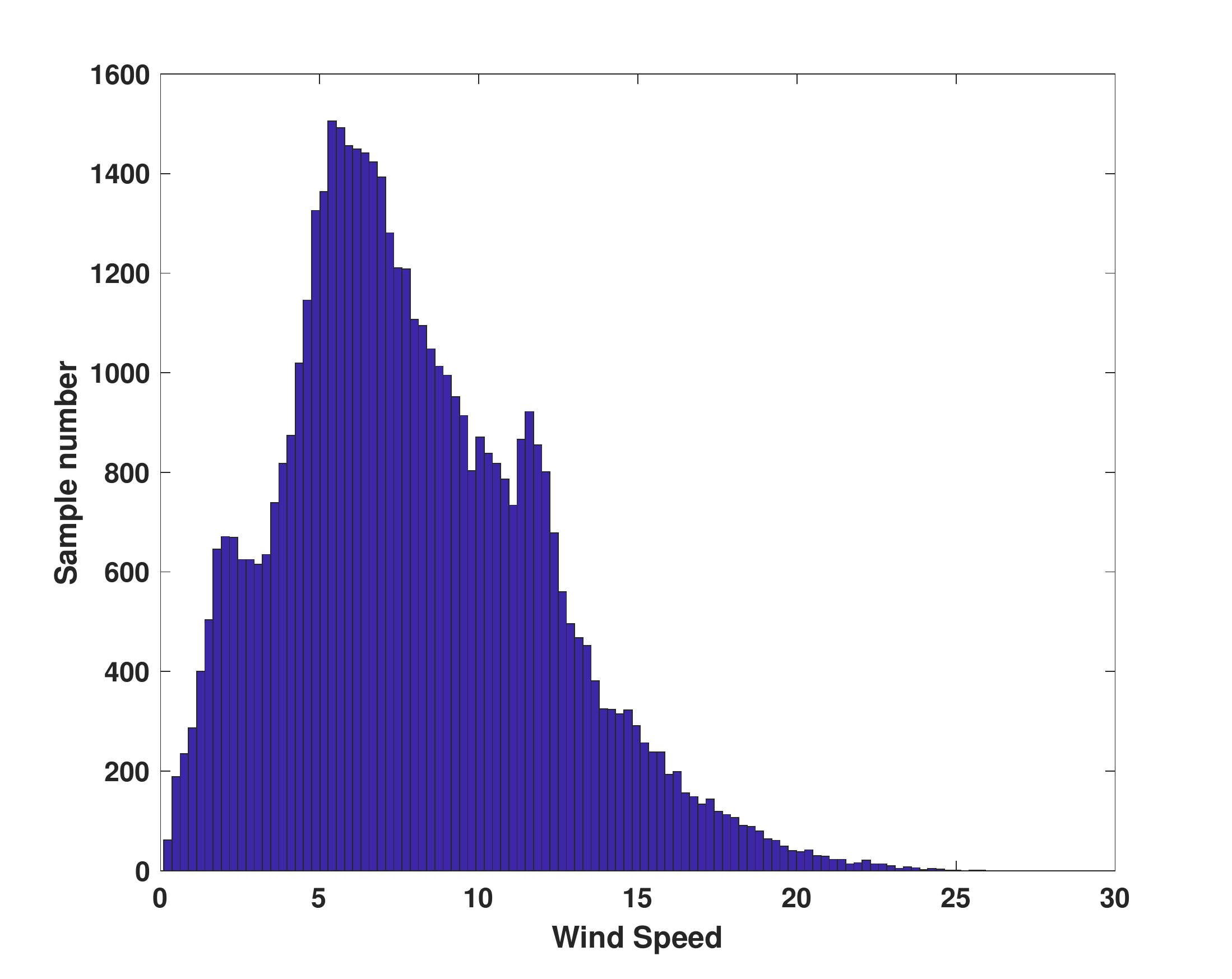}
   \caption{ Total distribution and frequency of the wind speed in Lillgrund Wind coastal site .  }
   \label{fig:total_distribution}
 
  \end{figure*}

The wind speed data collected from the Lillgrund wind farm (D3 wind turbine position can be seen in Figure~\ref{fig:positionD3}) consists of the period from July 2018 to July 2019 at a ten-minute resolution. Figures~\ref{fig:boxplot_speed} and~\ref{fig:total_distribution} show the distribution and frequency of the recorded wind speed in Lillgrund Wind farm during these 12 months. The distribution and frequency of the wind speed is strongly anisotropic\cite{soomere2003anisotropy}.

\begin{figure*}[tbp]
 \centering
  \includegraphics[width=0.85\textwidth]{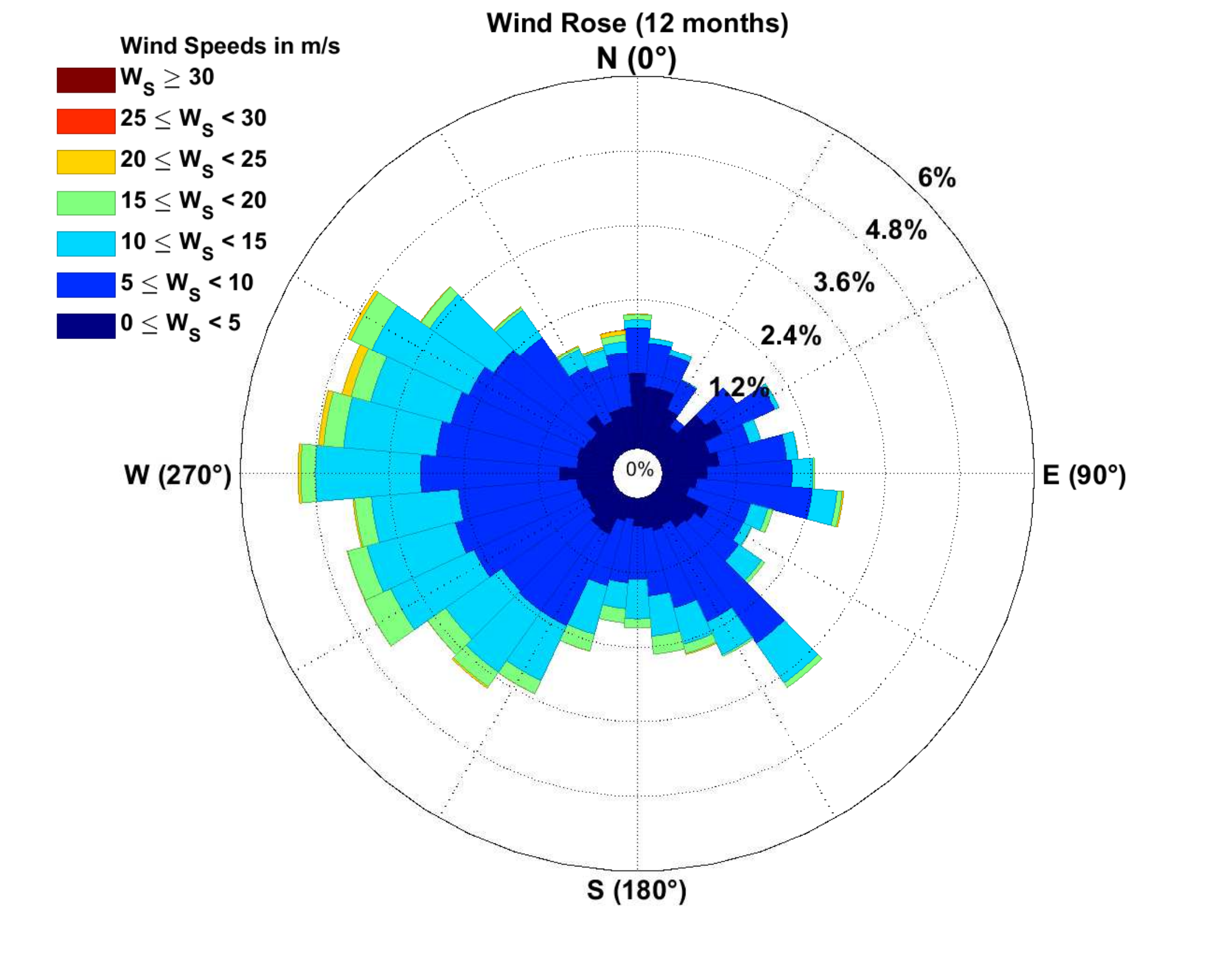}
   \caption{Wind rose: the speed and directional distribution of wind for the Lillgrund Wind coastal site. The dataset for generating this graph was obtained from \cite{jeppsson2008technical}   }
   \label{fig:wind_rose}
 
  \end{figure*}

Figure \ref{fig:wind_rose} shows that the dominating wind direction is south-west, and a secondary prevailing direction is south-east. 
However, there are also occasional North-west winds and sporadic north-east storms.

We use two horizons to predict wind speeds: ten minutes and one hour. 
 The wind speed data are randomly divided into three sets using blocks of indices including 80\% of the data is used as the training set, and the other 20\% is allocated as the test (10\%) and validation (10\%) set. And also We apply k-fold cross-validation for training the LSTM network in order to predict the time series data.

\section{Experiments and analysis}
\label{sec:Experiments}

For assessing the performance of the proposed CMAES-LSTM hybrid model,
we compare its performance with four well-known conventional forecasting techniques including FNN, ANFIS, PNN, NAR and one DNN forecasting model (the grid-search tuned LSTM network).  

In the first step of this study, a grid search algorithm is used to explore the search space of the hyper-parameters' impact on  LSTM Network performance. Conventionally, tuning hyper-parameters is done by hand and requires skilled practitioners\cite{smith2018disciplined}. Here the grid search is limited to tuning learning rate and batch size. Other parameters are fixed to allow the search to complete in reasonable time. 

Table~\ref{table:details_model} shows the final hyper-parameters of the models. 
We repeat each experiment ten times to ensure to allow for a reasonable sampling of each method's performance. 

\begin{figure}[t]
\centering
\subfloat[]{
\includegraphics[clip,width=0.5\columnwidth]{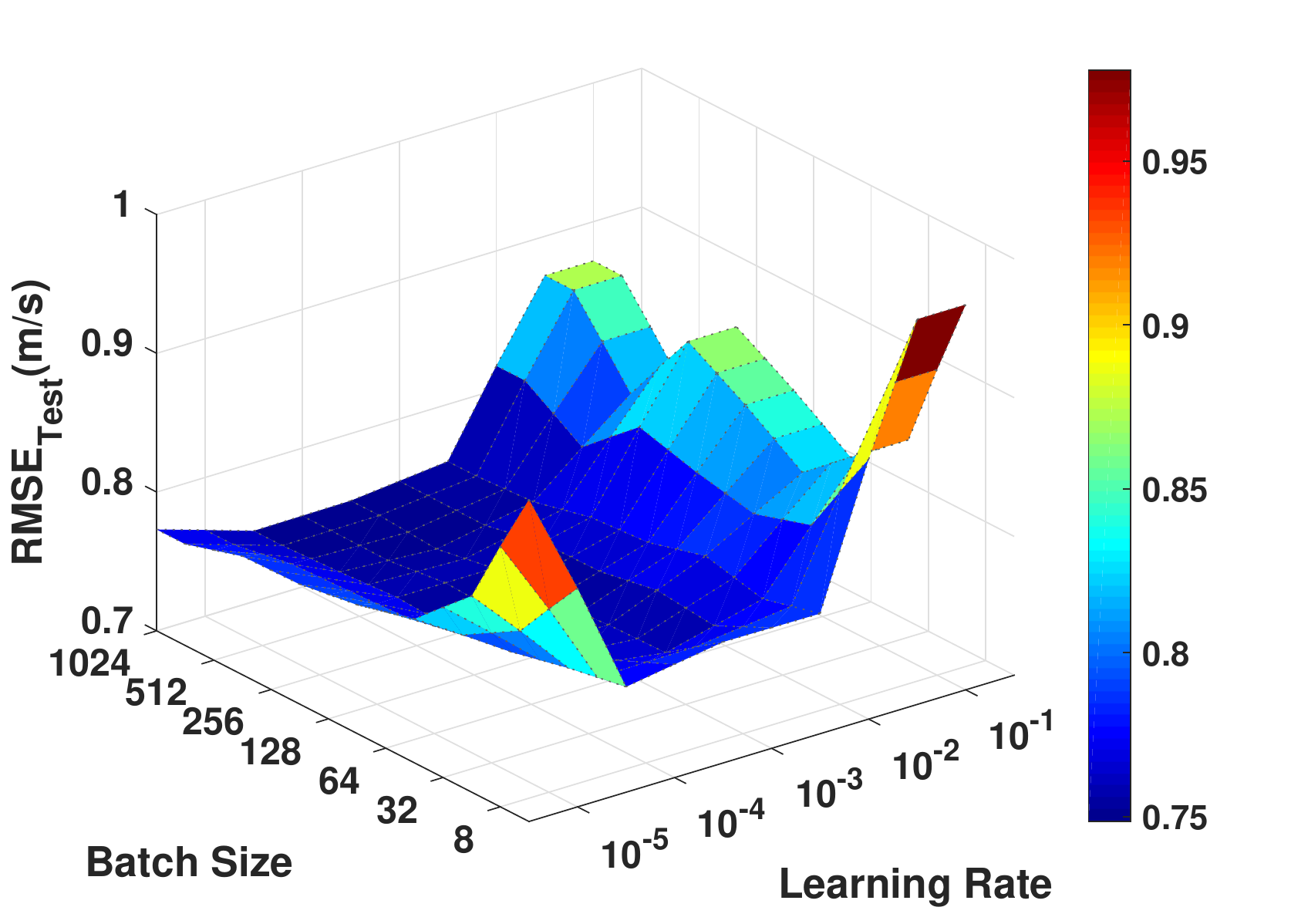}}
\subfloat[]{
\includegraphics[clip,width=0.5\columnwidth]{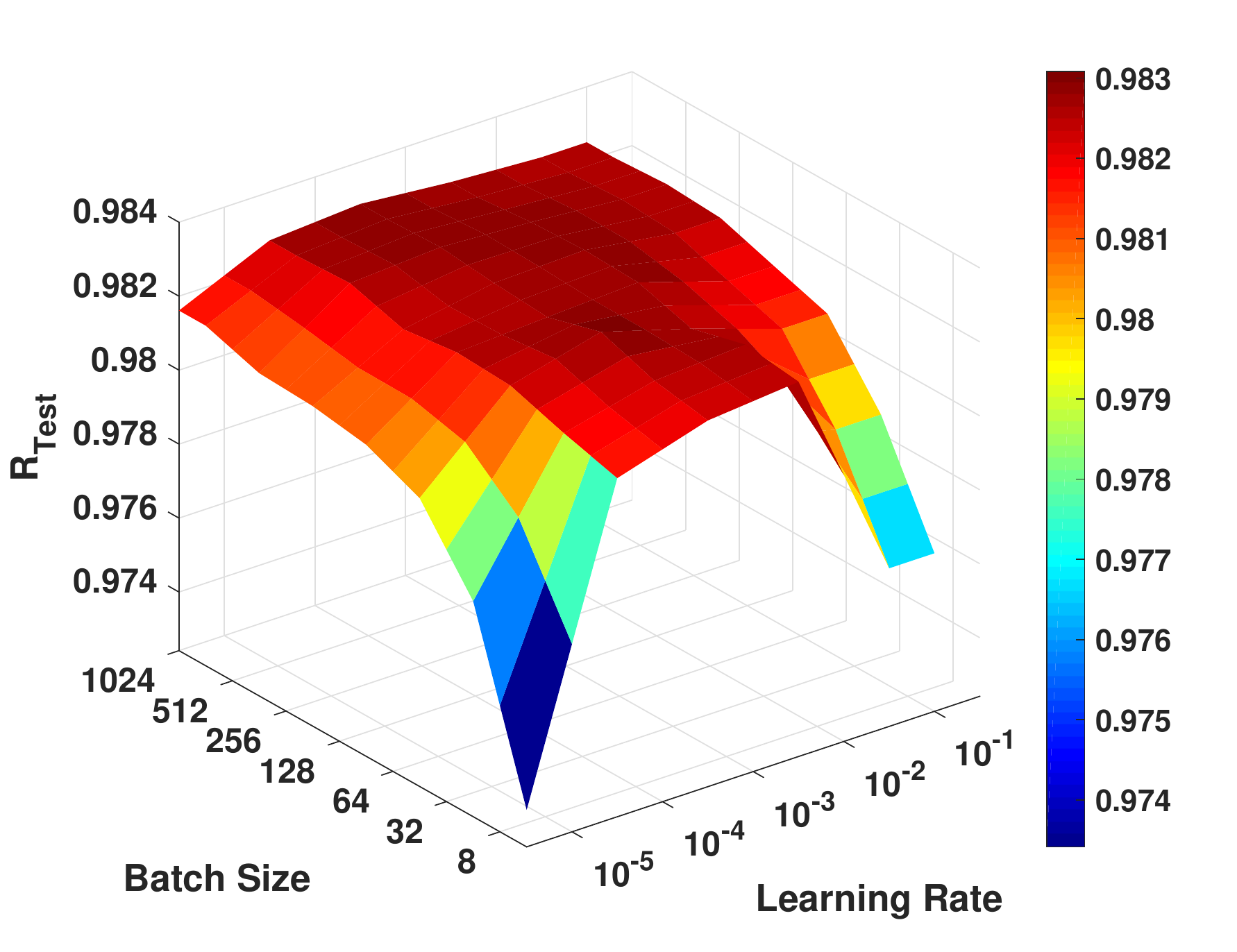}}\\
 \subfloat[]{
\includegraphics[clip,width=0.5\columnwidth]{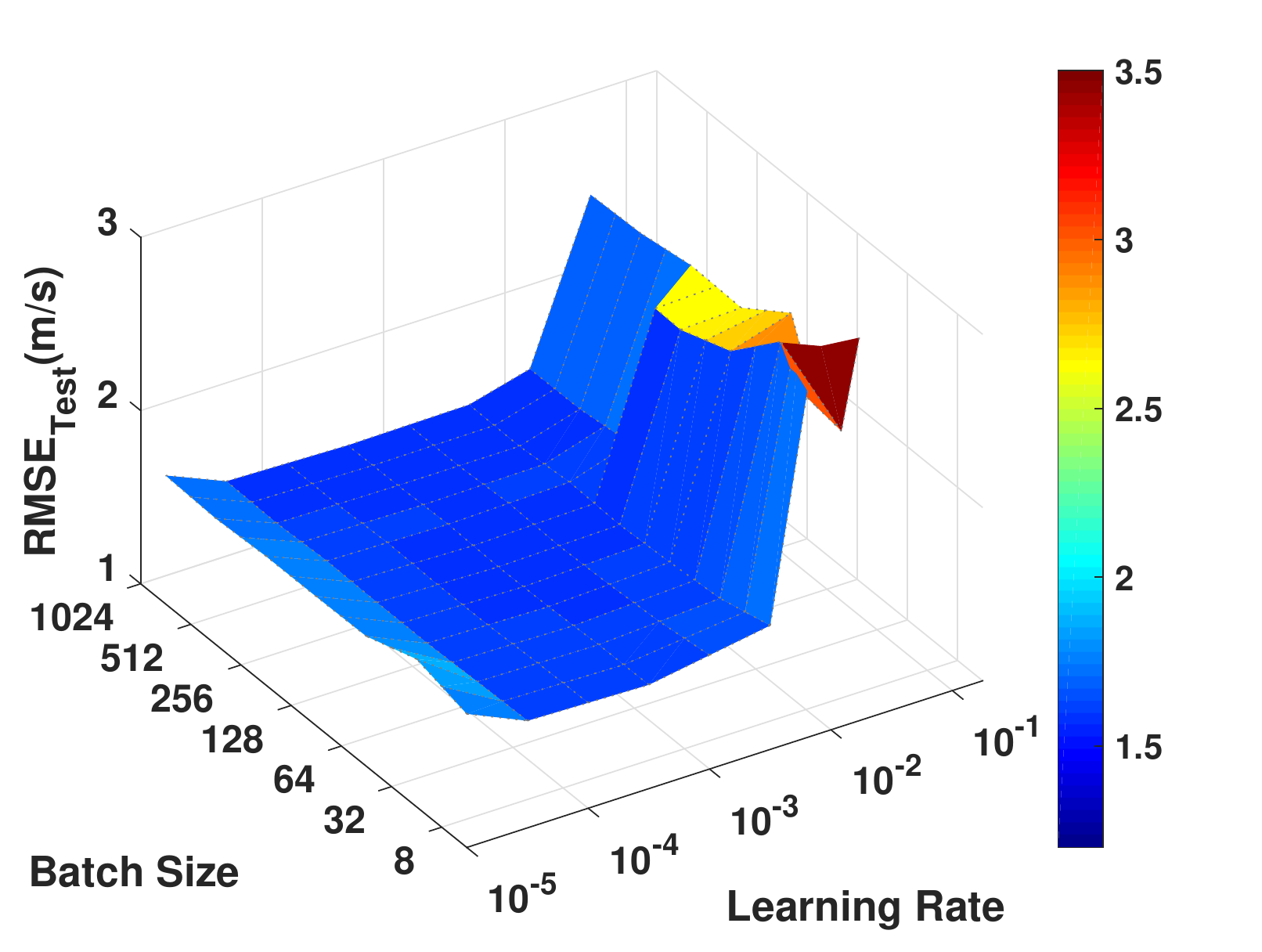}}
 \subfloat[]{
 \includegraphics[clip,width=0.5\columnwidth]{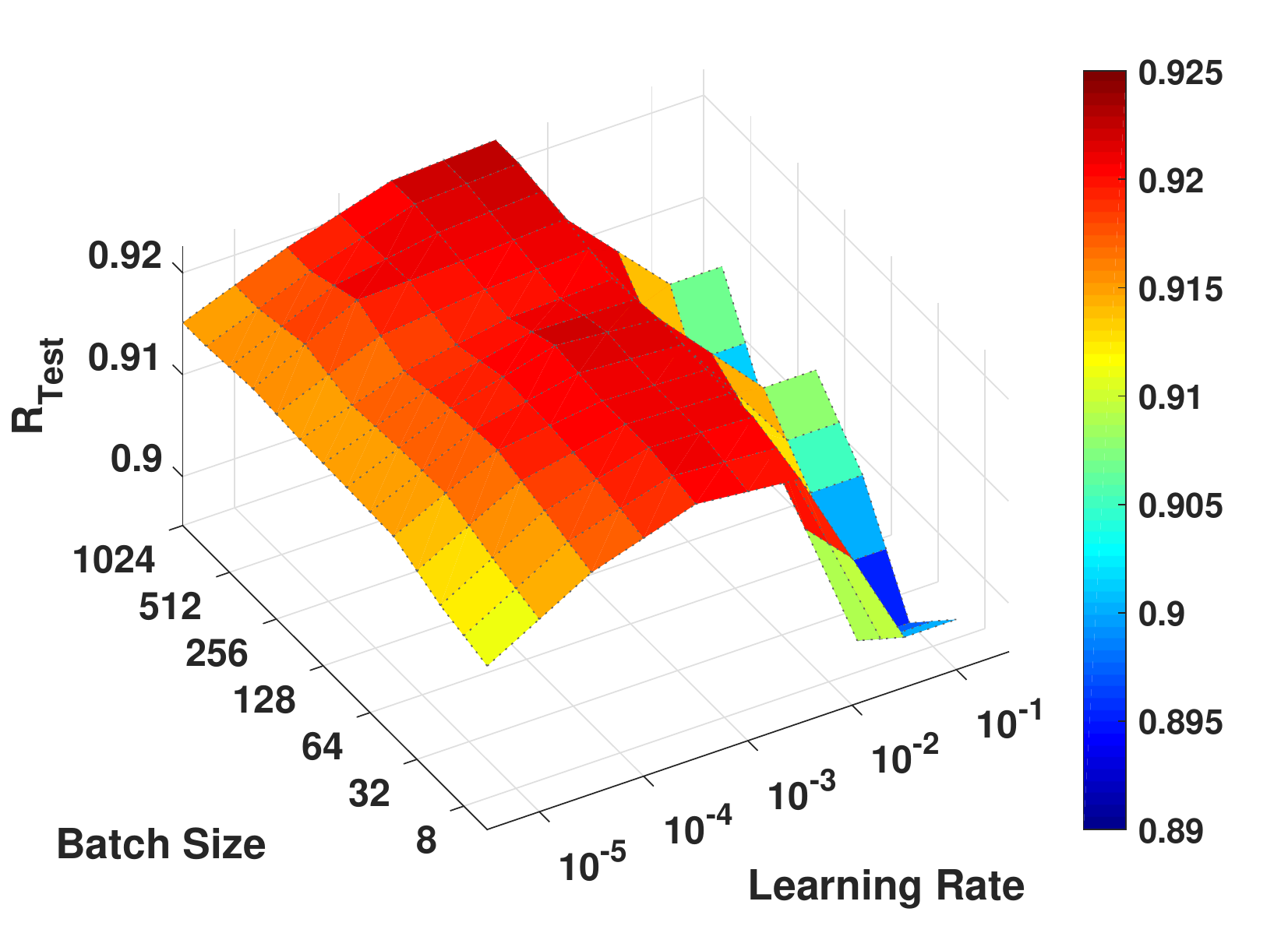}}
\caption{Hyper-parameters tuning of the applied LSTM network for forecasting the short-term wind speed .(a) the average of RMSE test-set (ten-minute ahead)  (b) the average of R-value test-set (ten-minute ahead), (c) the average of RMSE test-set (one-hour ahead)  (d) the average of R-value test-set (one-hour ahead)}%
\label{fig:3Dplot_gridsearch}
\end{figure}

Figure \ref{fig:3Dplot_gridsearch} demonstrates the forecasting results of tuning both batch size and learning rate in LSTM model performance for the two time-interval prediction dataset.  According to the observations, the minimum learning error  occurs where the learning rate is between $10^{-4}$ and $10^{-2}$.  The optimal size of batches is highly dependent on the selected learning rate values. 

\begin{figure*}[tbp]
 \subfloat[]{
\includegraphics[clip,width=\columnwidth]{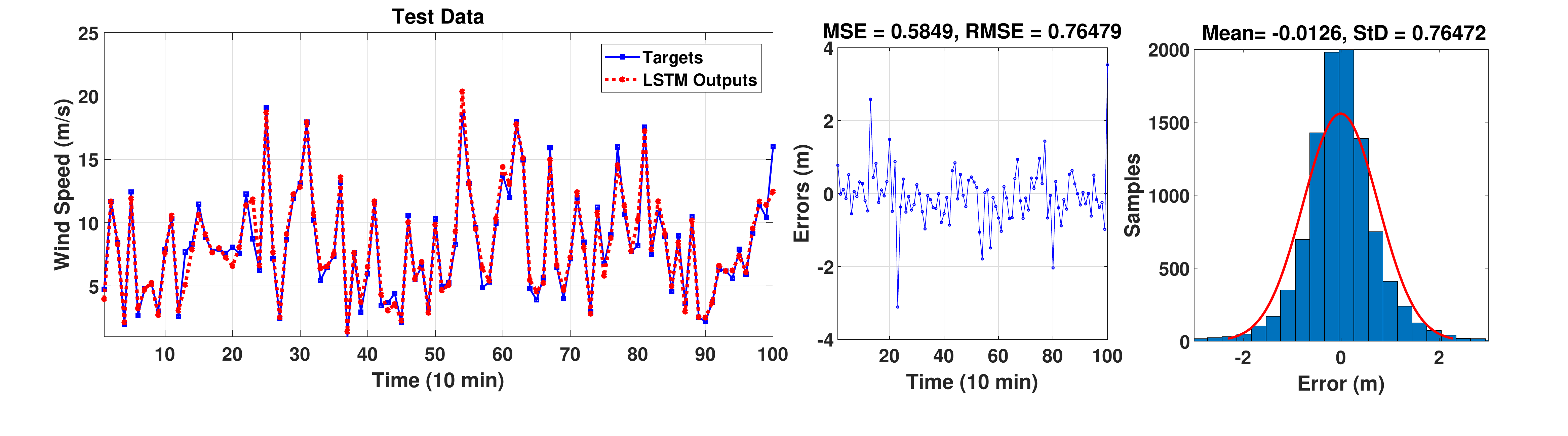}}\\
\subfloat[]{
\includegraphics[clip,width=\columnwidth]{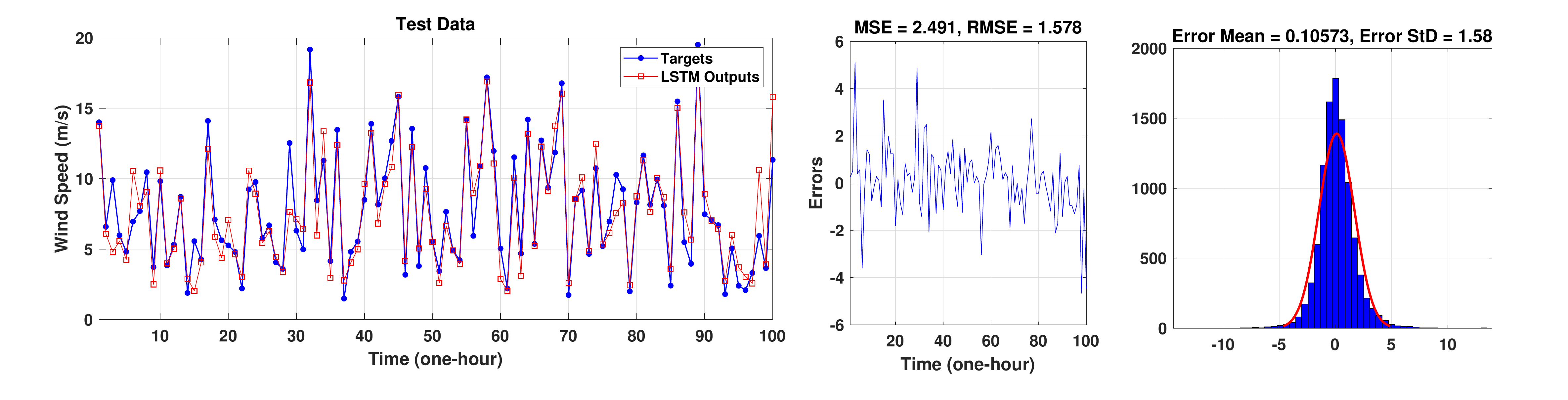}}\\
  \subfloat[]{
\includegraphics[clip,width=\columnwidth]{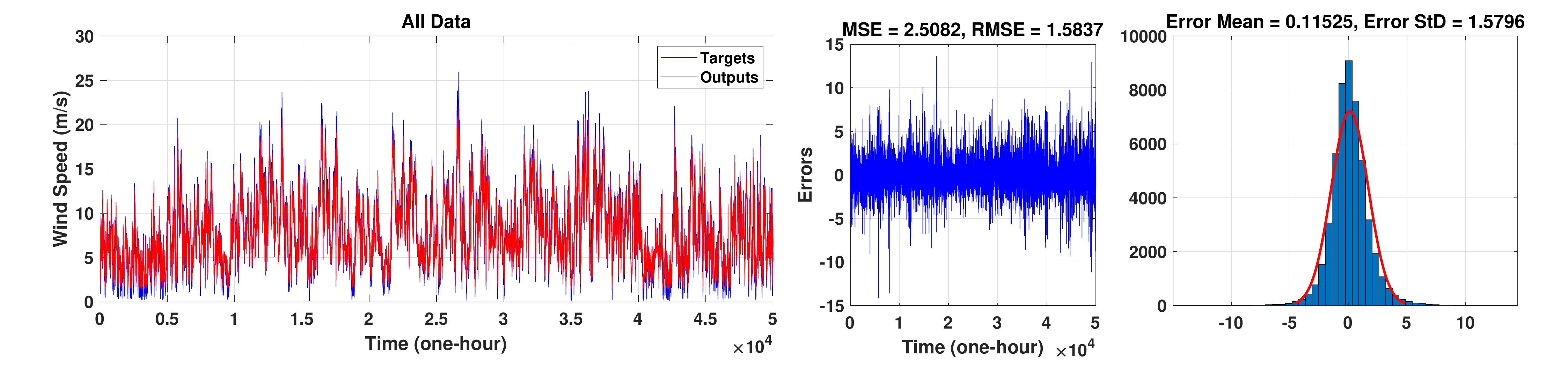}}\\
   \caption{ The wind speed forecasting results achieved by LSTM network with the best tuned hyper-parameters on (a) ten-minute ahead, (b) one-hour ahead (test data-set) and (c) one-hour ahead (all data-set)}
   \label{fig:error_plot}
 
  \end{figure*}

Figure \ref{fig:error_plot} shows the correlation between the original wind speed data with predicted ones for the ten-minute and one-hour forecast period.

The average errors of the testing model obtained by the best configuration of the grid-search tuned LSTM model are shown in Figure~\ref{fig:error_plot}. 

\begin{figure}[tbp]
 \centering
 
  \includegraphics[width=0.7\textwidth]{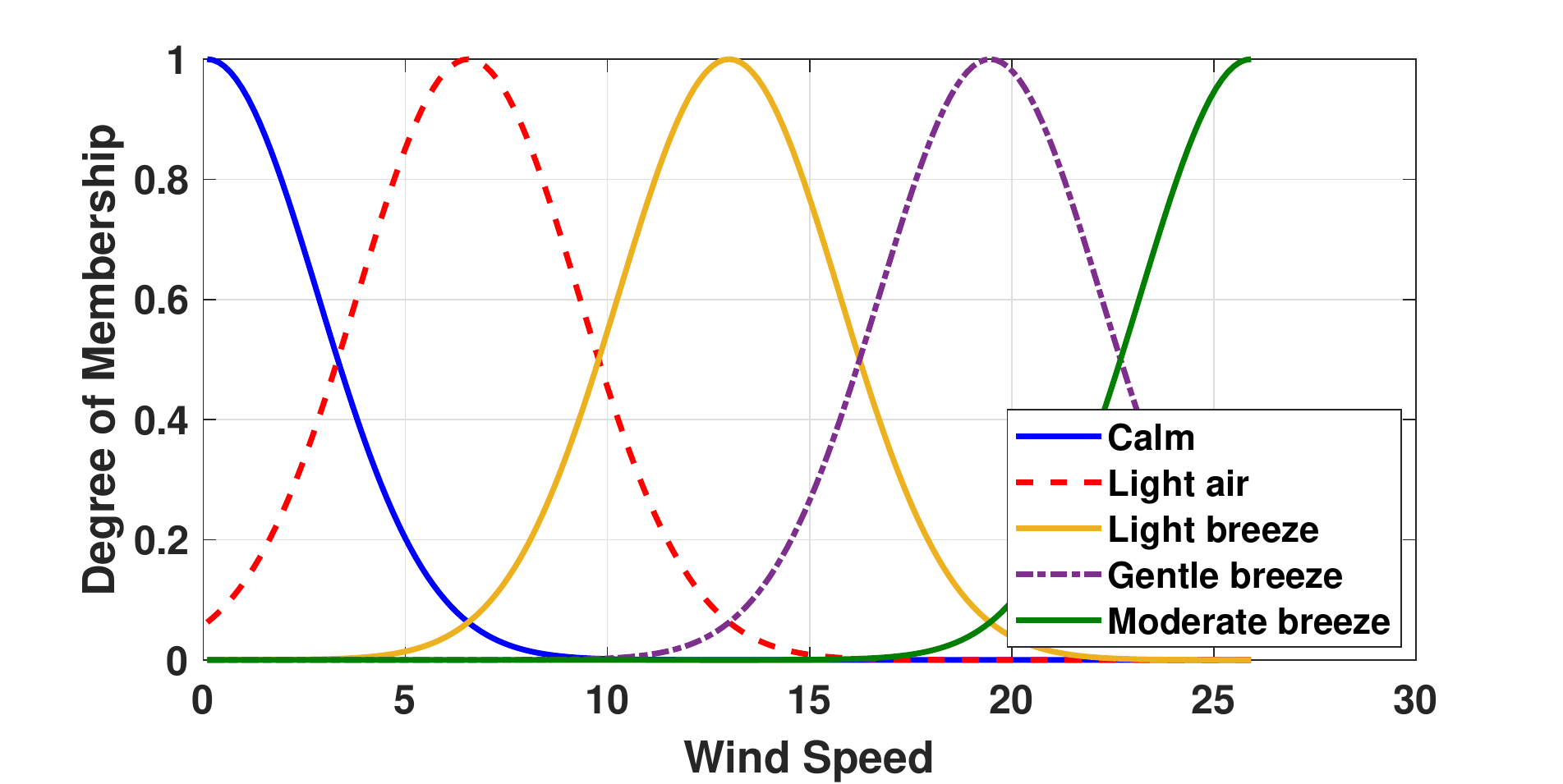}
   \caption{ fuzzy memberships applied for modelling the wind speed in ANFIS. }
   \label{fig:memfuzzy}
 
  \end{figure}

\begin{figure*}[tbp]
  \includegraphics[width=\textwidth]{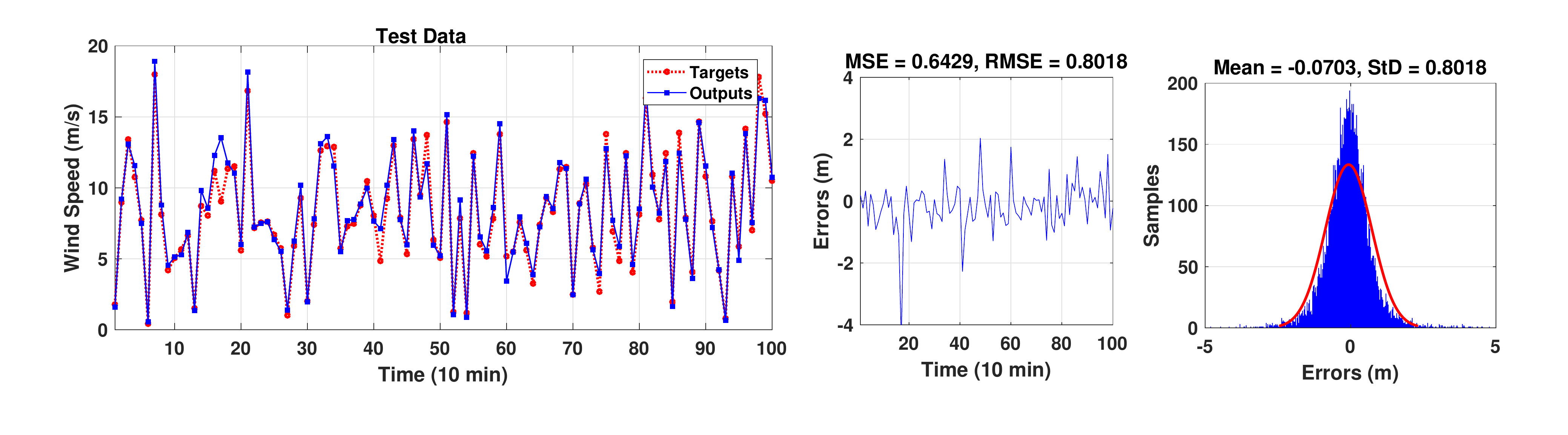}
   \caption{ The wind speed forecasting results achieved by ANFIS network with the best tuned hyper-parameters on ten-minute ahead  }
   \label{fig:abfis}
 
  \end{figure*}

One of the best forecasting models is adaptive neuro-fuzzy inference system (ANFIS) \cite{pousinho2011hybrid}. For modelling the wind speed by  proper membership functions, five Gaussian membership functions are defined to cover all range of the wind speed dataset (Figure \ref{fig:memfuzzy}). The performance of ANFIS is represented in Figure \ref{fig:abfis}.  We can see that the ANFIS estimation results are competitive. 

\begin{figure}[t]
\centering
\subfloat[]{
\includegraphics[clip,width=0.48\columnwidth]{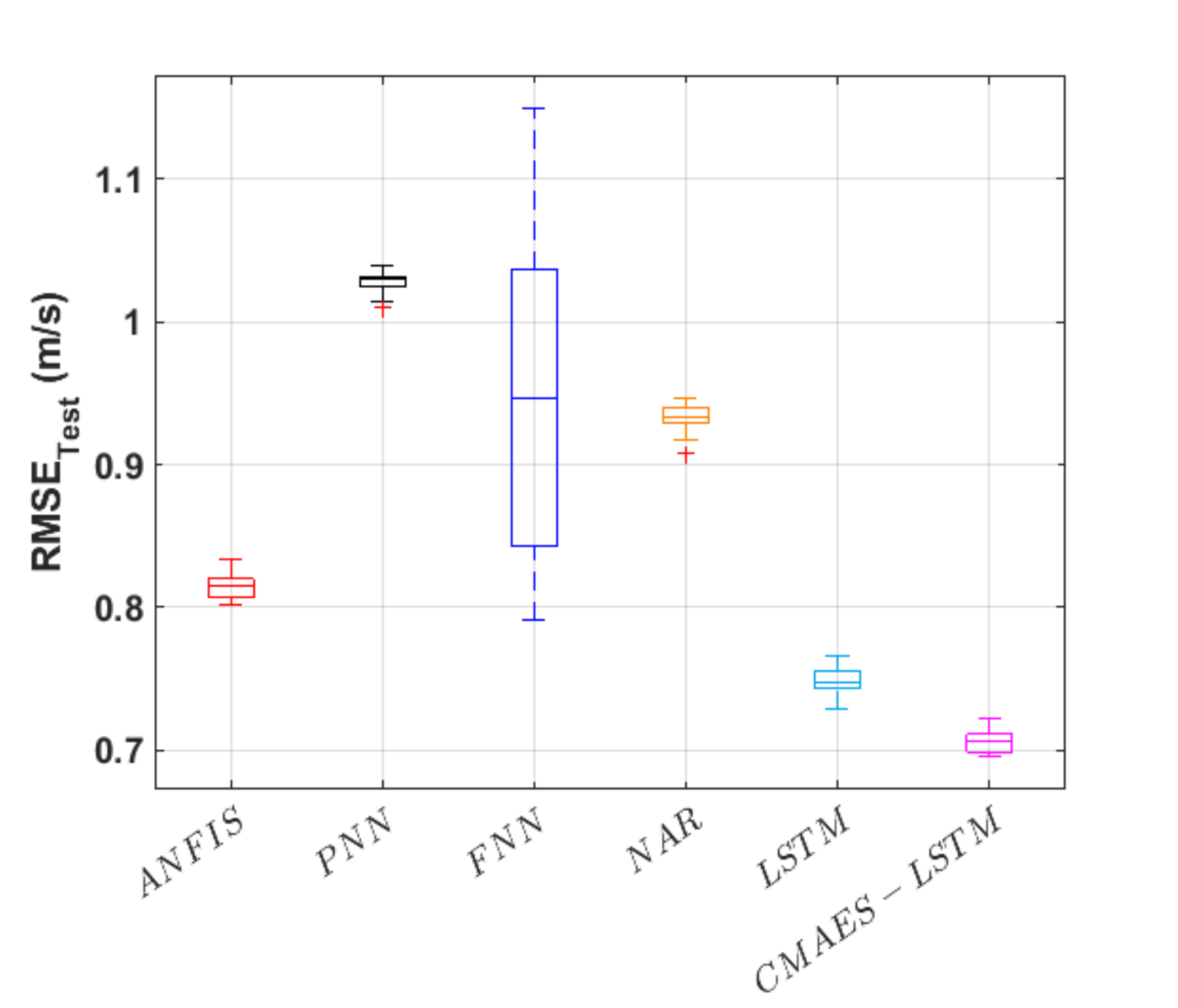}}
\subfloat[]{
\includegraphics[clip,width=0.48\columnwidth]{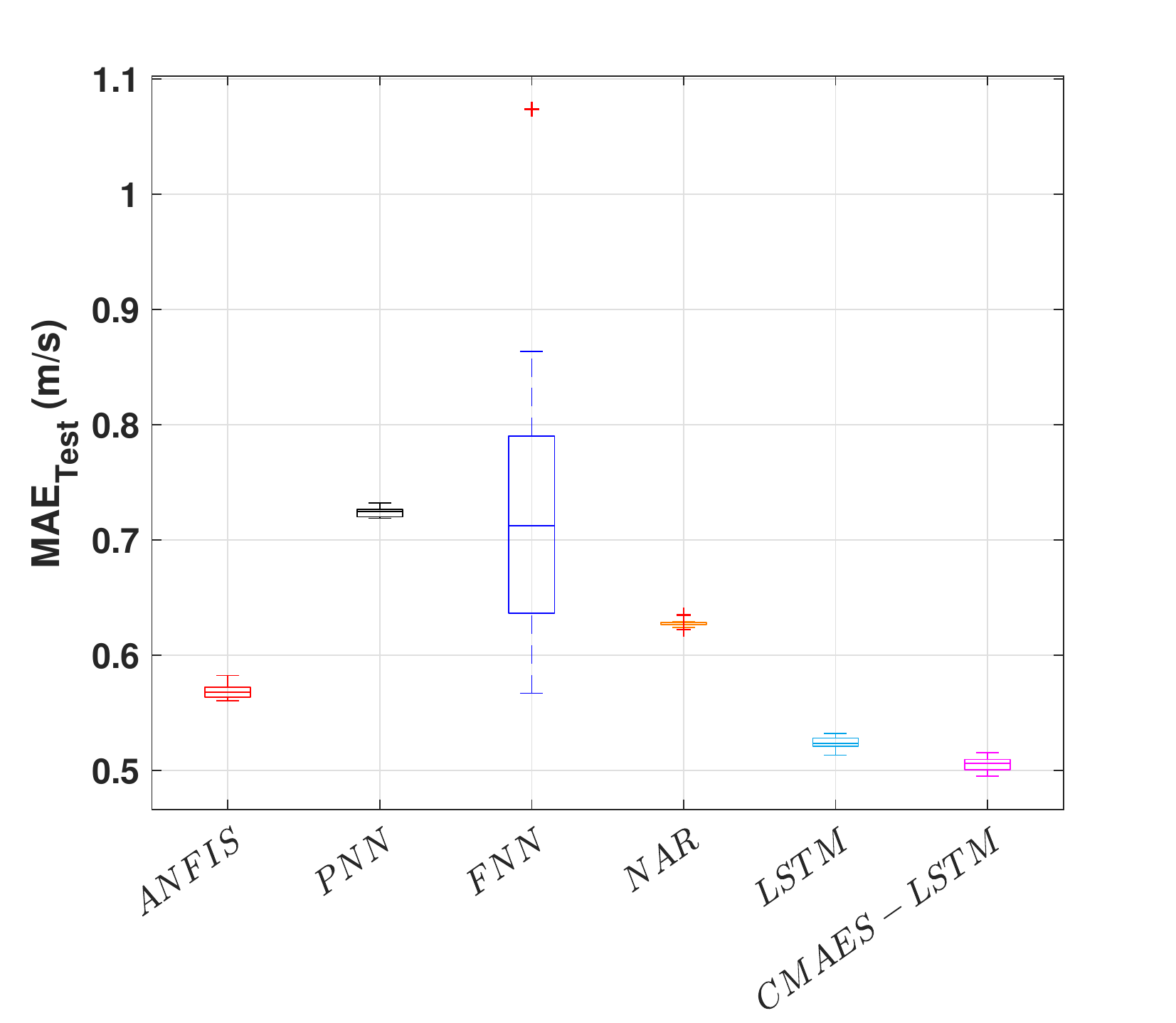}}\\
 \subfloat[]{
\includegraphics[clip,width=0.48\columnwidth]{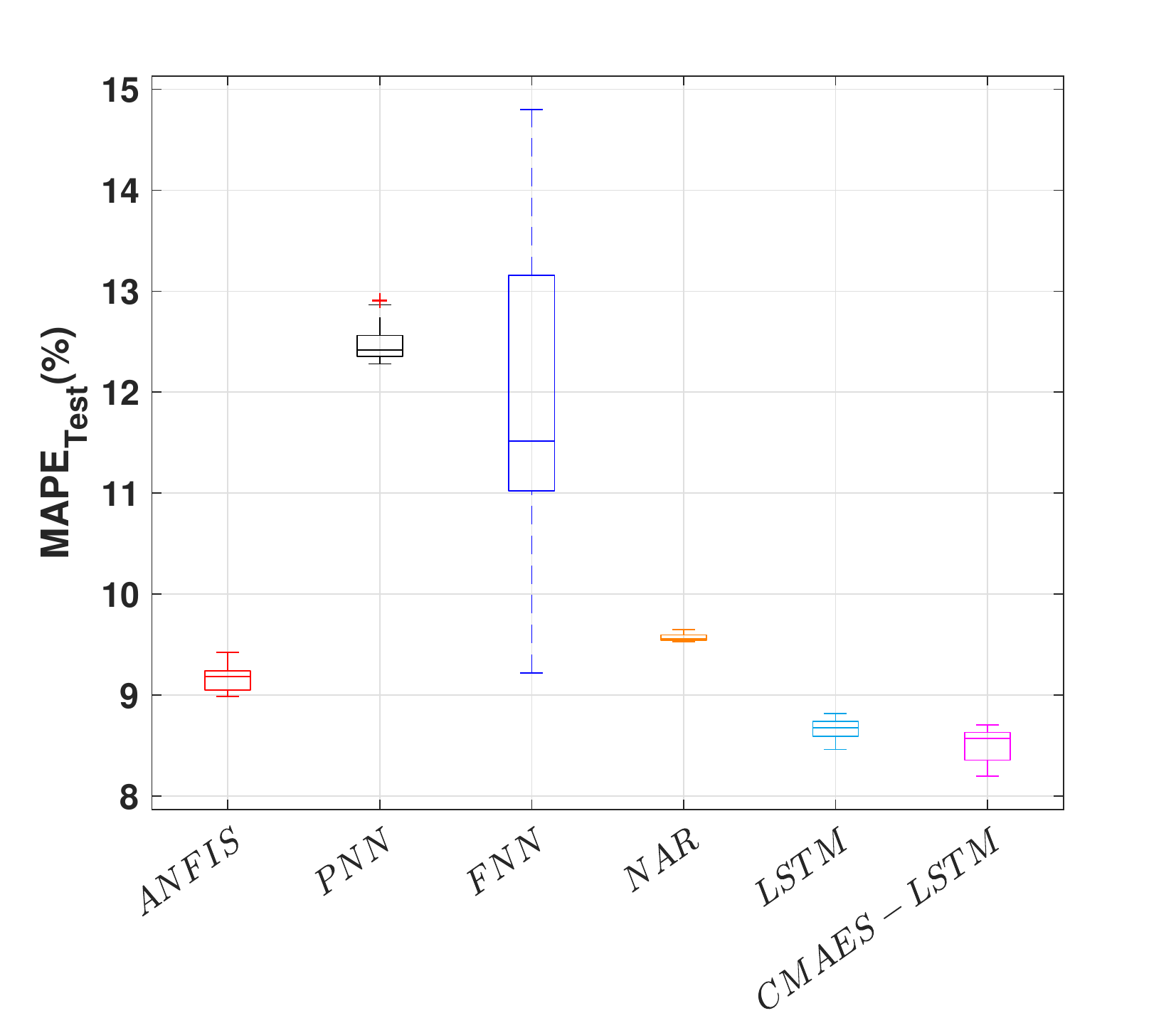}}
\subfloat[]{
\includegraphics[clip,width=0.48\columnwidth]{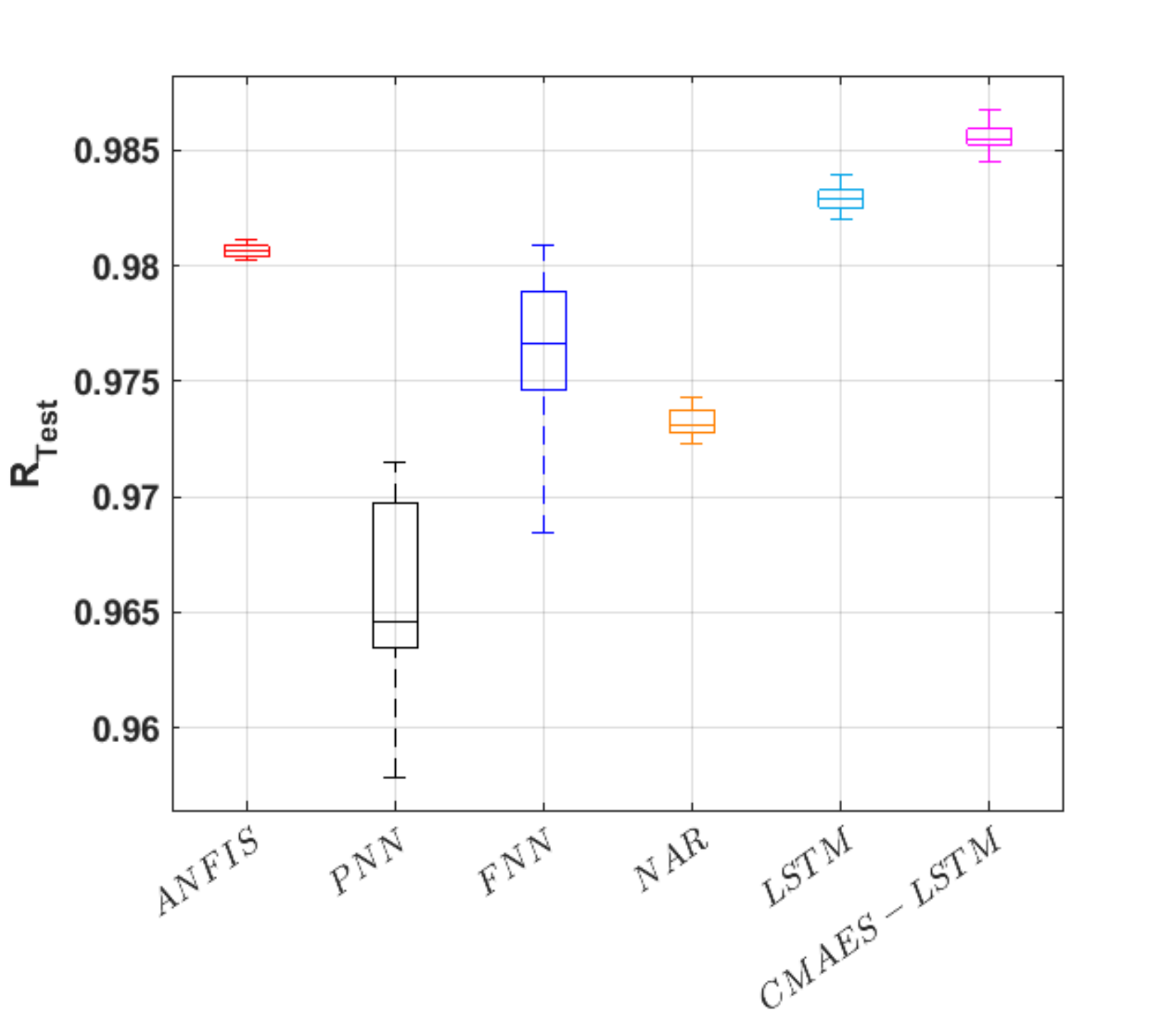}}\\
\caption{The performance of different forecasting models results of ten-minute ahead  (a) RMSE  (b) MAE (c) MAPE and (d) R-value .}%
\label{fig:boxplot_10min}
\end{figure}

Figure~\ref{fig:boxplot_10min} shows the results of the four performance indices applied in this work for short term wind speed forecasting (ten-minute ahead) received by five other models and the proposed hybrid model.
Concerning this experiment, the hybrid evolutionary model can outperform the other five competitors for short term wind speed forecasting with the minimum value of  RMSE as $0.695 (m/s)$, MAE as $0.495 (m/s)$, and MAPE as $8.2\%$ as well as the highest rate of R as $98.7$. 

Tables~\ref{table:stat_10min} and \ref{table:stat_onehour} summarise the statistical forecasting results for ten-minute and one-hour intervals. For both time intervals, our  CMAES-LSTM can accomplish better forecasting outcomes than other applied models. 

\begin{figure}[tbp]
 \centering
  \includegraphics[width=0.85\textwidth]{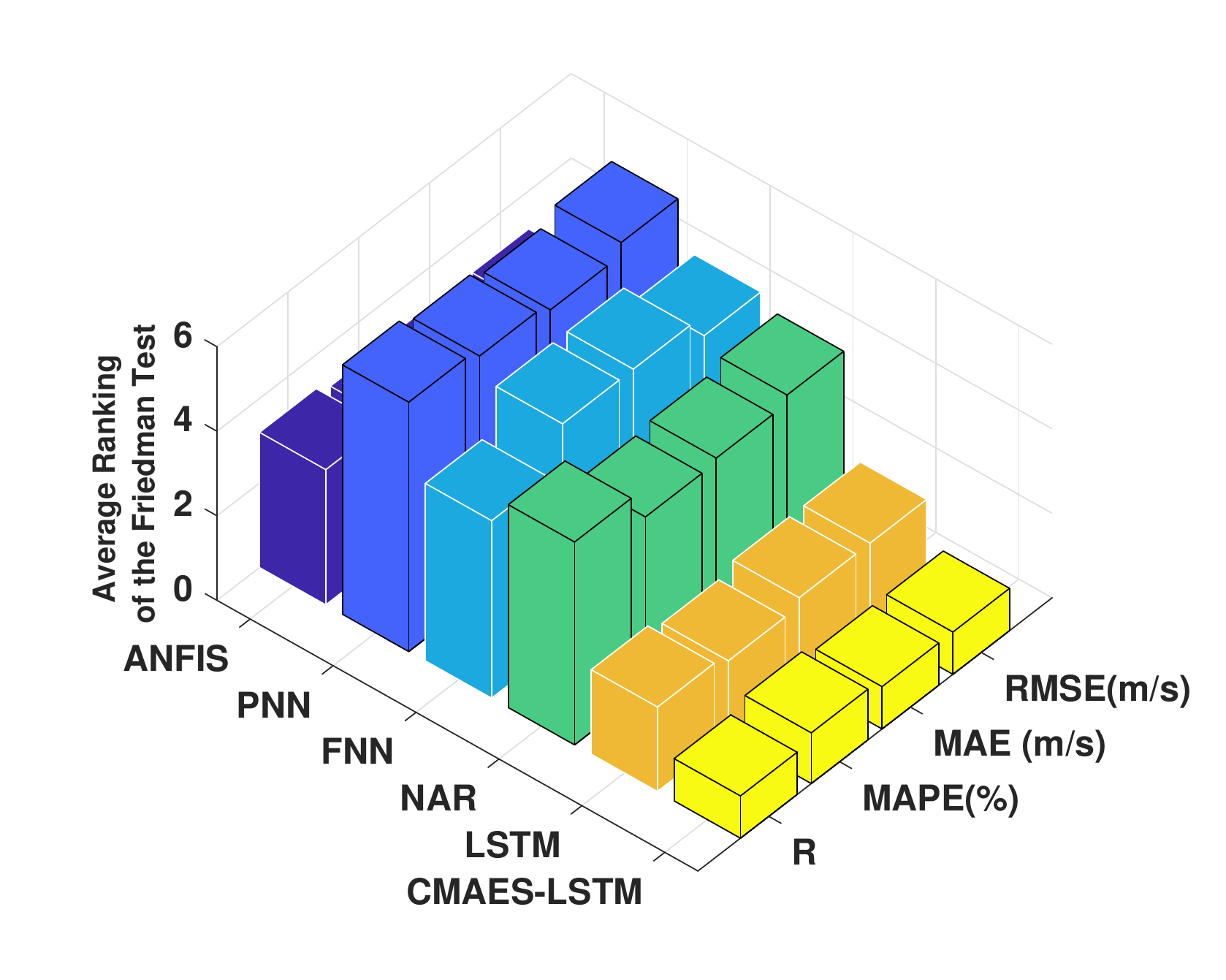} 
  \caption{Average ranking of the Friedman test for performance indices statistical tests achieved by various applied forecasting models (ten-minute ahead). }
   \label{fig:bar_friedman}
 
  \end{figure}

\begin{figure}[t]
\centering
 \subfloat[]{
\includegraphics[clip,width=0.35\columnwidth]{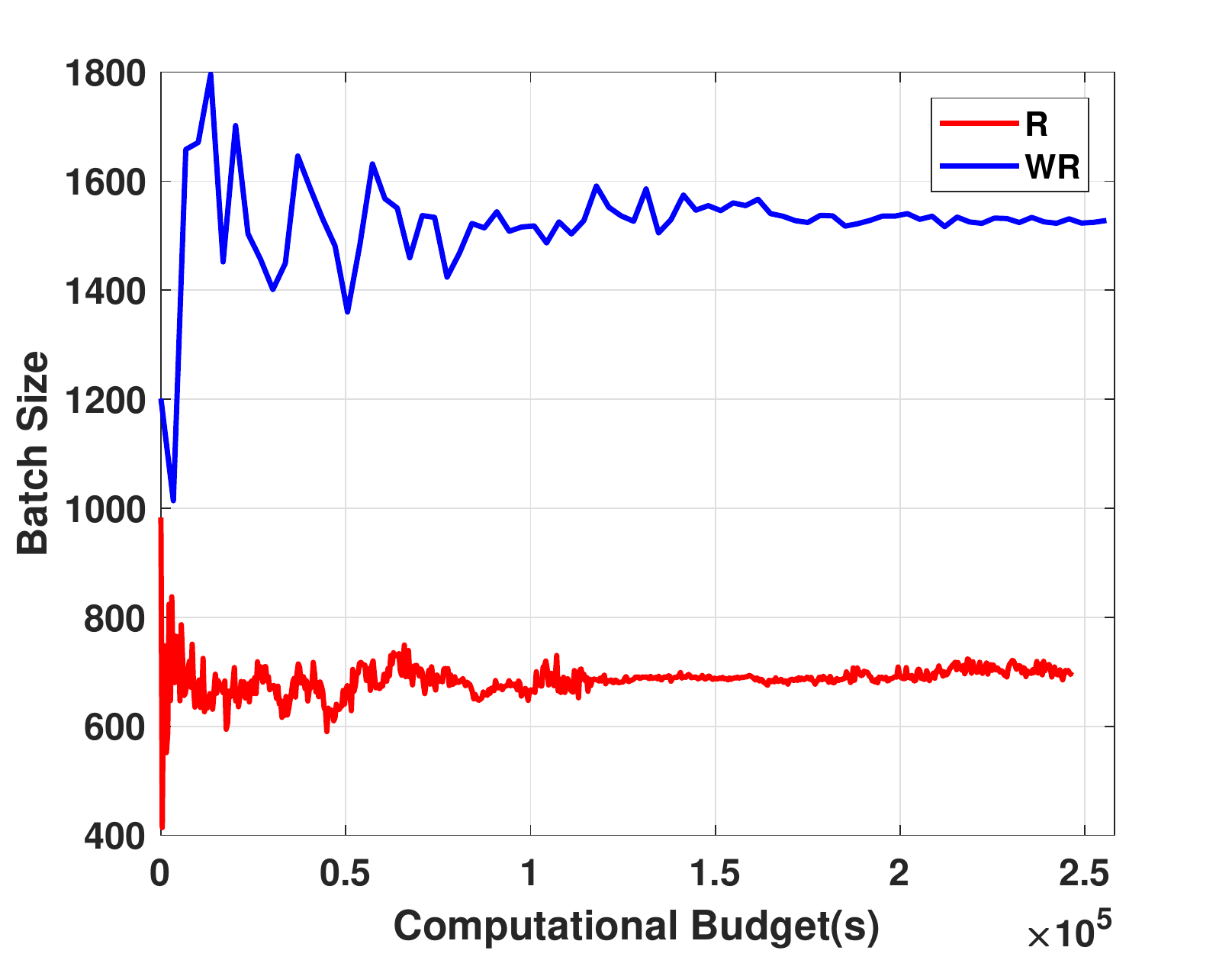}}
\subfloat[]{
\includegraphics[clip,width=0.35\columnwidth]{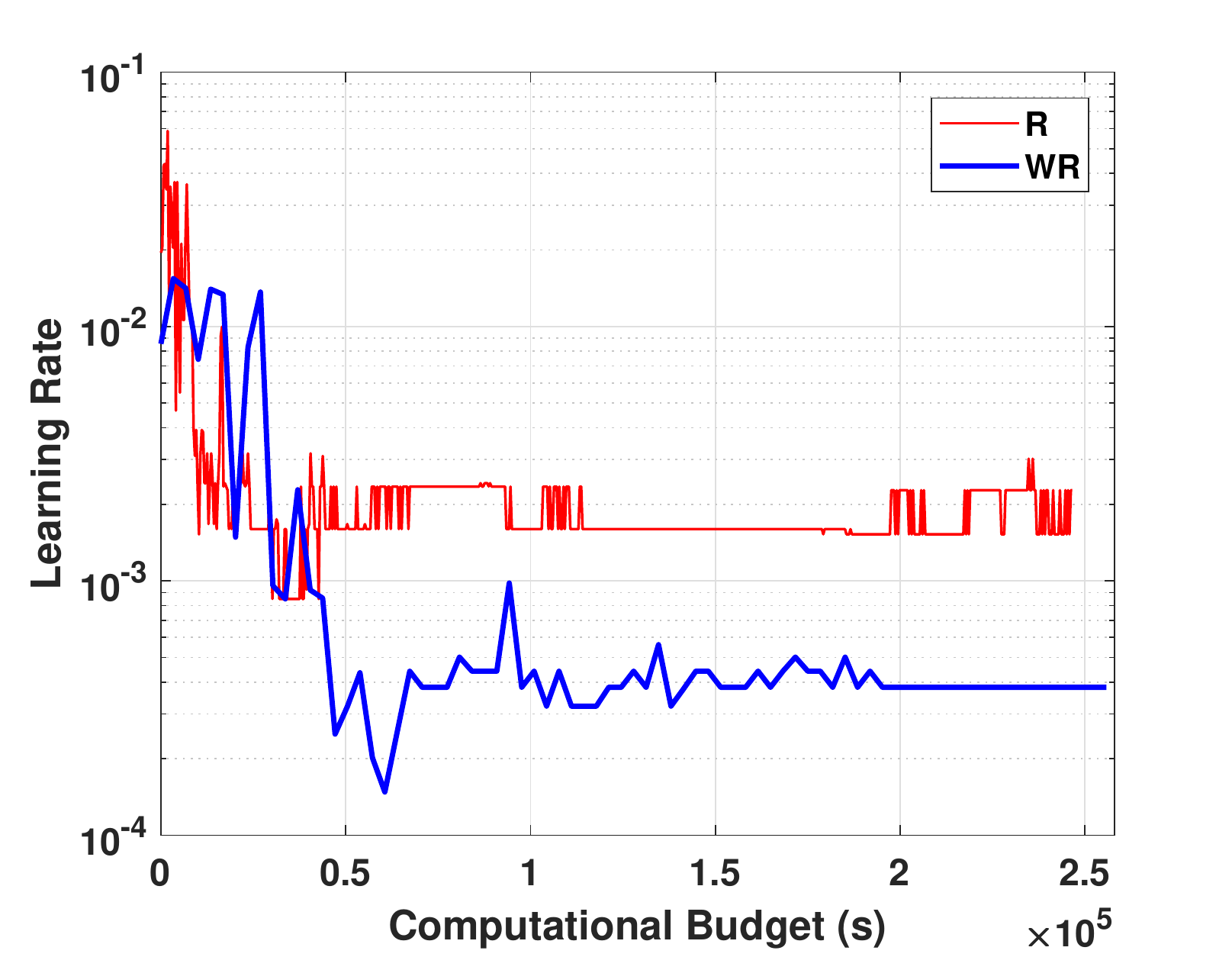}}
 \subfloat[]{
\includegraphics[clip,width=0.35\columnwidth]{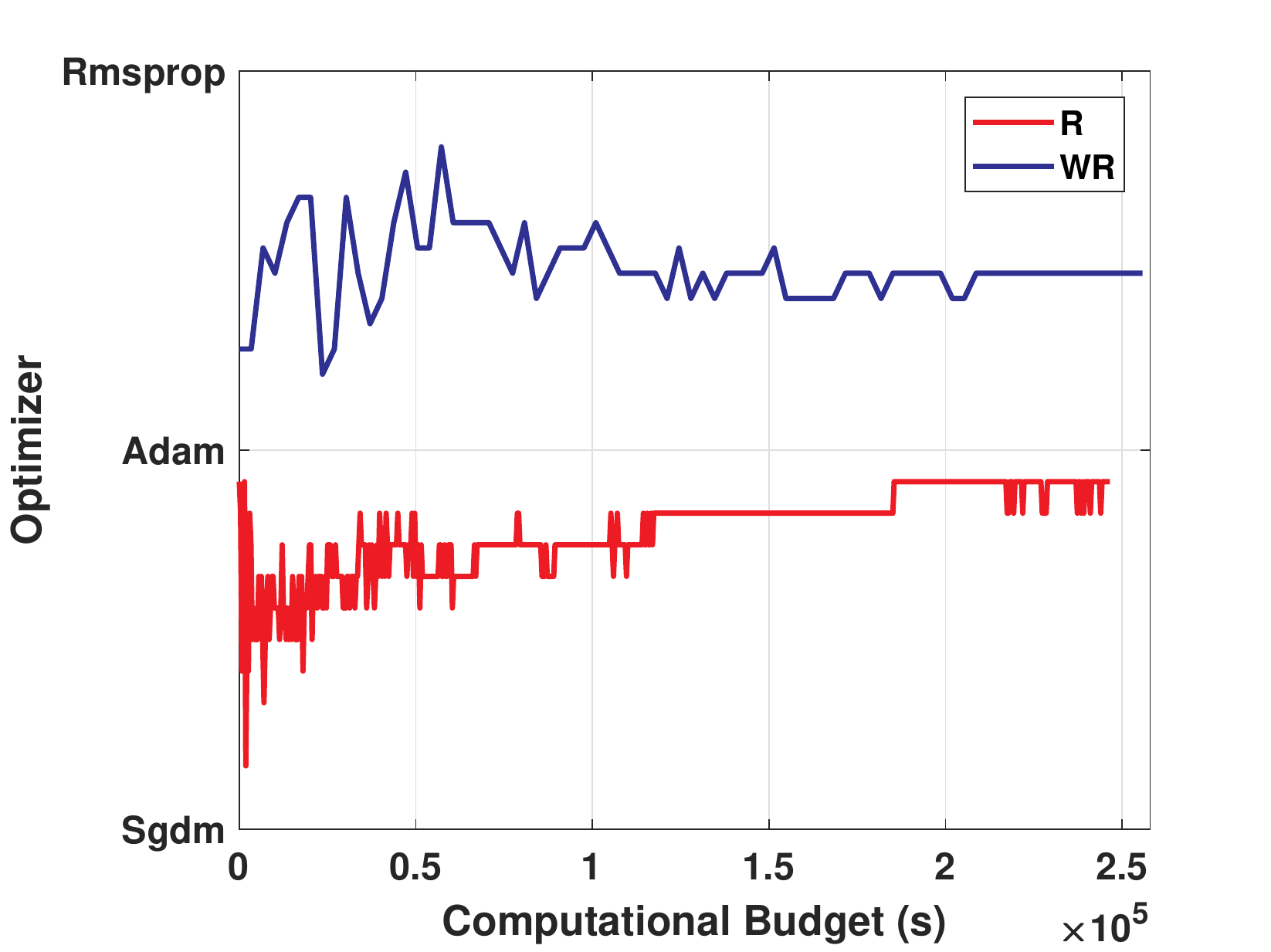}}\\
\subfloat[]{
\includegraphics[clip,width=0.35\columnwidth]{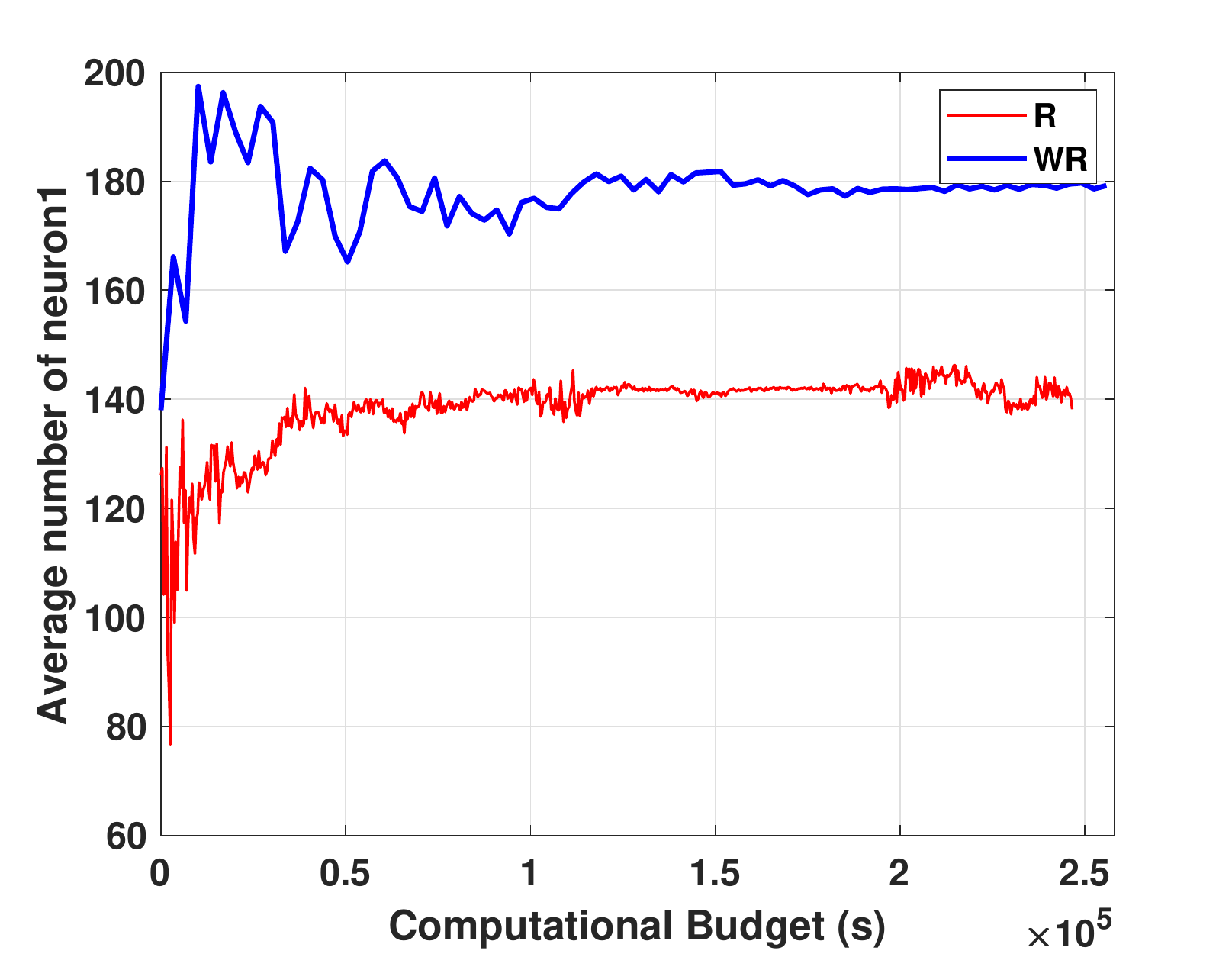}}
\subfloat[]{
\includegraphics[clip,width=0.35\columnwidth]{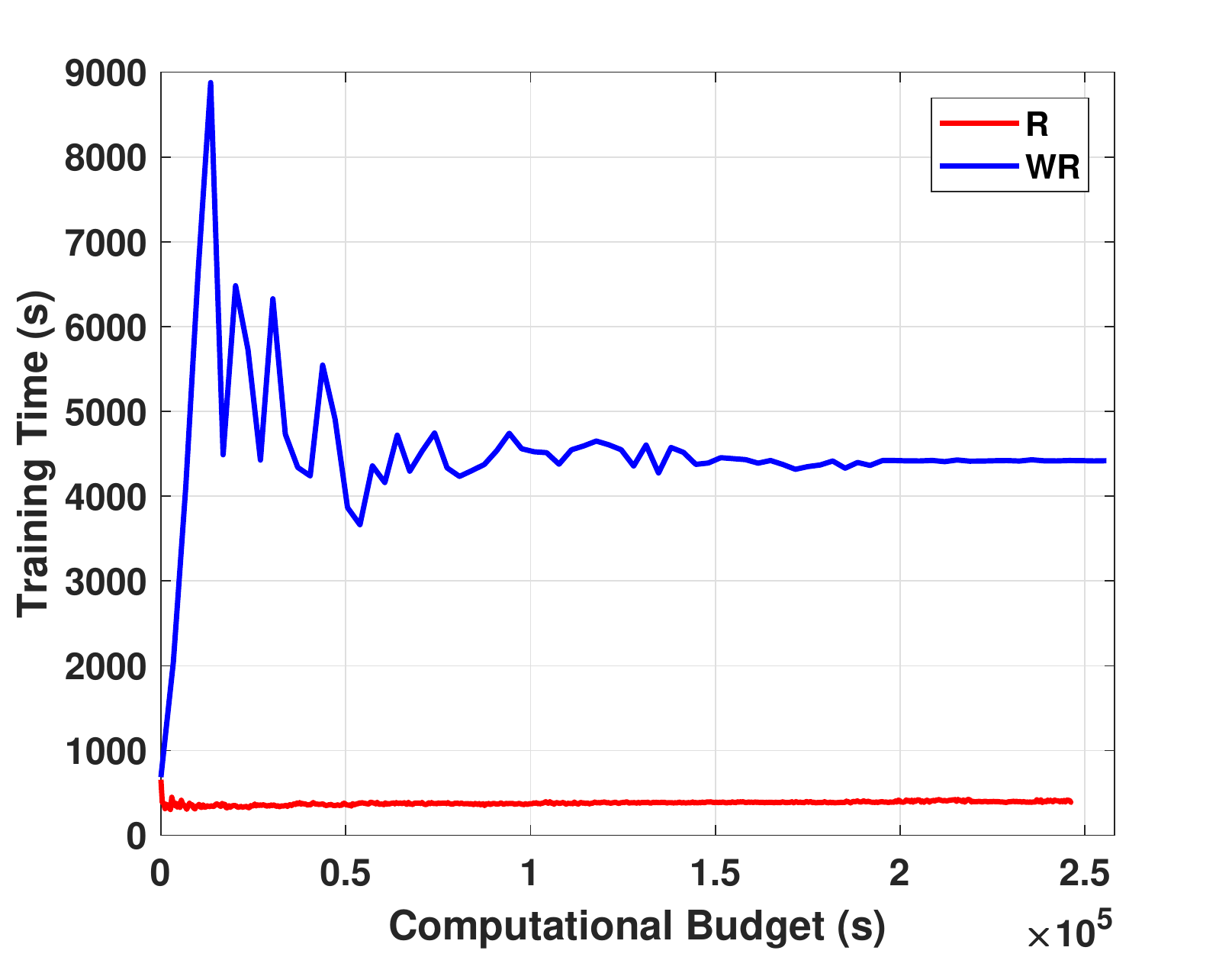}}

\caption{Comparison of CMAES performance for optimising the Hyper-parameters of the applied LSTM network for forecasting the short-term wind speed (10-minute ahead). For evaluating the performance of the LSTM model, two approaches are employed: applying a penalty for time consuming training runtime experiment (R) and without this penalty (WR). (a) the average of Batch Size (b) the average of learning rate, (c) Optimiser (Sgdm, Adam and Rmsprop), (d) average number of neurons in the first LSTM layer, (e) average of training runtime}%
\label{fig:comparison}
\end{figure}

In addition, in Figure~\ref{fig:bar_friedman} can be seen that CMAES-LSTM stands the first rank based on the Friedman statistic test with p-values less than $0.0001$, which signifies that the proposed forecasting method considerably performs better than other models. 
For evaluating the impact of the penalty factor on the Hybrid forecasting model performance, Figure \ref{fig:comparison} shows the comparison convergence of the hyper-parameters tuning process within (WR) and without (R) applying the penalty factor of the training runtime.  Interestingly, both cases are converged to the same learning rate at $10^{-3}$; however, in other parameters, the optimization results are different. It is noted that the whole allocated budget for both cases is the same, but the performance of CMAES-LSTM model with a training runtime penalty is better than another strategy.  


\begin{table}

\centering
\caption{Performance indices of forecasting outcomes achieved by different models on case
ten-minute ahead.}
\scalebox{0.7}{
\begin{tabular}{l|l|l|l|l|l|l|l|l|l|l|l}
\hline
\hlineB{4}
 & &    \multicolumn{2}{ c }\textbf{MSE(m/s)} &  \multicolumn{2}{ c }\textbf{RMSE(m/s)}    & \multicolumn{2}{ c }\textbf{MAE(m/s)}& \multicolumn{2}{ c }\textbf{MAPE(\%)} & \multicolumn{1}{ c }\textbf{R}          \\  \hline
 
\textbf{Model}&  & \textbf{Train}    & \textbf{Test}     & \textbf{Train}    & \textbf{Test}     & \textbf{Train}    & \textbf{Test}     & \textbf{Train}    & \textbf{Test}     & \textbf{Train}    & \textbf{Test}     \\  \hline
      & Mean & 6.60E-01 & 6.64E-01 & 8.13E-01 & 8.15E-01 & 5.69E-01 & 5.69E-01 & 9.19E+00 & 9.17E+00 & 9.81E-01 & 9.81E-01 \\
  \textbf{ANFIS}         & Min  & 6.53E-01 & 6.43E-01 & 8.08E-01 & 8.02E-01 & 5.66E-01 & 5.61E-01 & 9.12E+00 & 8.99E+00 & 9.80E-01 & 9.80E-01 \\
           & Max  & 6.71E-01 & 6.96E-01 & 8.19E-01 & 8.34E-01 & 5.75E-01 & 5.82E-01 & 9.27E+00 & 9.42E+00 & 9.81E-01 & 9.81E-01 \\
           & Std  & 4.58E-03 & 1.55E-02 & 2.82E-03 & 9.47E-03 & 2.49E-03 & 7.07E-03 & 4.83E-02 & 1.39E-01 & 1.99E-04 & 3.23E-04 \\
      \hlineB{4}     
       & Mean & 1.06E+00 & 1.06E+00 & 1.03E+00 & 1.03E+00 & 7.25E-01 & 7.24E-01 & 1.25E+01 & 1.25E+01 & 9.67E-01 & 9.67E-01 \\
  \textbf{PNN}          & Min  & 1.05E+00 & 1.02E+00 & 1.03E+00 & 1.01E+00 & 7.22E-01 & 7.19E-01 & 1.24E+01 & 1.23E+01 & 9.66E-01 & 9.66E-01 \\
           & Max  & 1.08E+00 & 1.08E+00 & 1.04E+00 & 1.04E+00 & 7.28E-01 & 7.32E-01 & 1.26E+01 & 1.29E+01 & 9.69E-01 & 9.70E-01 \\
           & Std  & 7.64E-03 & 1.78E-02 & 3.70E-03 & 8.70E-03 & 2.04E-03 & 4.69E-03 & 7.40E-02 & 2.22E-01 & 2.40E-03 & 3.10E-03 \\
           \hlineB{4}
      & Mean & 9.73E-01 & 9.56E-01 & 9.78E-01 & 9.68E-01 & 7.24E-01 & 7.19E-01 & 1.22E+01 & 1.22E+01 & 9.76E-01 & 9.77E-01 \\
   \textbf{FFNN}         & Min  & 6.26E-01 & 6.28E-01 & 7.91E-01 & 7.92E-01 & 5.61E-01 & 5.61E-01 & 9.11E+00 & 9.22E+00 & 9.68E-01 & 9.68E-01 \\
           & Max  & 1.25E+00 & 1.24E+00 & 1.12E+00 & 1.11E+00 & 8.56E-01 & 8.53E-01 & 1.45E+01 & 1.44E+01 & 9.81E-01 & 9.82E-01 \\
           & Std  & 2.77E-01 & 2.84E-01 & 1.38E-01 & 1.42E-01 & 1.32E-01 & 1.34E-01 & 2.30E+00 & 2.24E+00 & 5.04E-03 & 4.84E-03 \\
           \hlineB{4}
        & Mean & 8.69E-01 & 8.69E-01 & 9.32E-01 & 9.32E-01 & 6.27E-01 & 6.28E-01 & 9.57E+00 & 9.51E+00 & 9.73E-01 & 9.73E-01 \\
  \textbf{NAR}         & Min  & 8.56E-01 & 8.47E-01 & 9.25E-01 & 9.20E-01 & 6.24E-01 & 6.21E-01 & 9.54E+00 & 9.30E+00 & 9.73E-01 & 9.72E-01 \\
           & Max  & 8.78E-01 & 8.87E-01 & 9.37E-01 & 9.42E-01 & 6.29E-01 & 6.32E-01 & 9.61E+00 & 9.60E+00 & 9.74E-01 & 9.74E-01 \\
           & Std  & 8.83E-03 & 1.87E-02 & 4.73E-03 & 9.98E-03 & 2.65E-03 & 4.44E-03 & 3.66E-02 & 8.70E-02 & 3.39E-04 & 5.43E-04 \\
           \hlineB{4}
     & Mean & 5.64E-01 & 5.60E-01 & 7.51E-01 & 7.48E-01 & 5.23E-01 & 5.24E-01 & 8.65E+00 & 8.66E+00 & 9.83E-01 & 9.83E-01 \\
  \textbf{LSTM}           & Min  & 5.55E-01 & 5.31E-01 & 7.45E-01 & 7.29E-01 & 5.19E-01 & 5.13E-01 & 8.50E+00 & 8.46E+00 & 9.83E-01 & 9.82E-01 \\
           & Max  & 5.70E-01 & 5.76E-01 & 7.55E-01 & 7.59E-01 & 5.26E-01 & 5.29E-01 & 8.74E+00 & 8.77E+00 & 9.83E-01 & 9.83E-01 \\
           & Std  & 5.35E-03 & 1.61E-02 & 3.56E-03 & 1.08E-02 & 2.64E-03 & 5.60E-03 & 8.93E-02 & 1.10E-01 & 1.42E-04 & 5.48E-04 \\
           \hlineB{4}
& Mean & 5.59E-01 & 5.00E-01 & 7.61E-01 & 7.27E-01 & 5.20E-01 & 5.07E-01 & 8.70E+00 & 8.57E+00 & 9.83E-01 & 9.85E-01 \\
 \textbf{CMAES-LSTM   }        & Min  & 5.46E-01 & 4.83E-01 & 7.39E-01 & 7.19E-01 & 5.12E-01 & 4.95E-01 & 8.58E+00 & 8.20E+00 & 9.83E-01 & 9.84E-01 \\
           & Max  & 5.65E-01 & 5.22E-01 & 7.52E-01 & 7.30E-01 & 5.25E-01 & 5.16E-01 & 8.76E+00 & 8.71E+00 & 9.84E-01 & 9.87E-01 \\
           & Std  & 4.45E-03 & 1.41E-02 & 3.56E-03 & 1.28E-02 & 3.64E-03 & 7.60E-03 & 5.93E-02 & 1.60E-01 & 4.42E-04 & 2.48E-04\\
\hlineB{4}
\end{tabular}
}
\label{table:stat_10min}
\end{table}
\begin{table}[]

\centering
\caption{Performance indices of forecasting outcomes achieved by different models on the case of one-hour ahead.}

\scalebox{0.7}{
\begin{tabular}{l|l|l|l|l|l|l|l|l|l|l|l}

\hline
\hlineB{4}
 & &    \multicolumn{2}{ c }\textbf{MSE(m/s)} &  \multicolumn{2}{ c }\textbf{RMSE(m/s)}    & \multicolumn{2}{ c }\textbf{MAE(m/s)}& \multicolumn{2}{ c }\textbf{MAPE(\%)} & \multicolumn{1}{ c }\textbf{R}          \\  \hline
 
\textbf{Model}&  & \textbf{Train}    & \textbf{Test}     & \textbf{Train}    & \textbf{Test}     & \textbf{Train}    & \textbf{Test}     & \textbf{Train}    & \textbf{Test}     & \textbf{Train}    & \textbf{Test}     \\  \hline
ANFIS      & Mean & 2.60E+00 & 2.59E+00 & 1.61E+00 & 1.61E+00 & 1.16E+00 & 1.16E+00 & 2.05E+01 & 2.05E+01 & 9.19E-01 & 9.19E-01 \\
           & Min  & 2.58E+00 & 2.49E+00 & 1.61E+00 & 1.58E+00 & 1.16E+00 & 1.15E+00 & 2.04E+01 & 2.02E+01 & 9.18E-01 & 9.16E-01 \\
           & Max  & 2.62E+00 & 2.66E+00 & 1.62E+00 & 1.63E+00 & 1.17E+00 & 1.17E+00 & 2.06E+01 & 2.08E+01 & 9.20E-01 & 9.21E-01 \\
           & Std  & 1.55E-02 & 6.62E-02 & 4.80E-03 & 2.05E-02 & 2.54E-03 & 1.09E-02 & 7.18E-02 & 2.59E-01 & 4.99E-04 & 1.94E-03 \\
           \hlineB{4}
PNN        & Mean & 3.93E+00 & 3.91E+00 & 1.98E+00 & 1.98E+00 & 1.48E+00 & 1.48E+00 & 3.01E+01 & 3.03E+01 & 8.73E-01 & 8.73E-01 \\
           & Min  & 3.90E+00 & 3.85E+00 & 1.97E+00 & 1.96E+00 & 1.48E+00 & 1.47E+00 & 2.98E+01 & 2.97E+01 & 8.70E-01 & 8.72E-01 \\
           & Max  & 3.95E+00 & 3.96E+00 & 1.99E+00 & 1.99E+00 & 1.49E+00 & 1.49E+00 & 3.03E+01 & 3.06E+01 & 8.75E-01 & 8.74E-01 \\
           & Std  & 2.02E-02 & 4.72E-02 & 5.10E-03 & 1.19E-02 & 3.63E-03 & 8.71E-03 & 1.52E-01 & 3.80E-01 & 1.90E-03 & 9.00E-04 \\
           \hlineB{4}
FFNN       & Mean & 3.39E+00 & 3.41E+00 & 1.82E+00 & 1.82E+00 & 1.36E+00 & 1.36E+00 & 2.61E+01 & 2.62E+01 & 9.15E-01 & 8.95E-01 \\
           & Min  & 2.65E+00 & 2.59E+00 & 1.63E+00 & 1.61E+00 & 1.17E+00 & 1.14E+00 & 2.04E+01 & 1.97E+01 & 9.05E-01 & 8.95E-01 \\
           & Max  & 4.65E+00 & 4.66E+00 & 2.12E+00 & 2.12E+00 & 1.61E+00 & 1.62E+00 & 3.20E+01 & 3.24E+01 & 9.20E-01 & 8.99E-01 \\
           & Std  & 1.26E+00 & 1.26E+00 & 2.97E-01 & 2.96E-01 & 2.56E-01 & 2.59E-01 & 5.99E+00 & 6.22E+00 & 5.03E-03 & 4.70E-03 \\
           \hlineB{4}
NAR        & Mean & 3.55E+00 & 3.57E+00 & 1.88E+00 & 1.89E+00 & 1.44E+00 & 1.44E+00 & 3.49E+01 & 3.59E+01 & 8.95E-01 & 8.95E-01 \\
           & Min  & 3.49E+00 & 3.46E+00 & 1.87E+00 & 1.86E+00 & 1.43E+00 & 1.41E+00 & 3.46E+01 & 3.46E+01 & 8.95E-01 & 8.93E-01 \\
           & Max  & 3.59E+00 & 3.65E+00 & 1.90E+00 & 1.91E+00 & 1.45E+00 & 1.46E+00 & 3.52E+01 & 3.67E+01 & 8.96E-01 & 8.97E-01 \\
           & Std  & 3.83E-02 & 8.08E-02 & 1.01E-02 & 2.14E-02 & 8.42E-03 & 1.89E-02 & 2.20E-01 & 8.66E-01 & 5.59E-04 & 1.49E-03 \\
           \hlineB{4}
LSTM       & Mean & 2.50E+00 & 2.49E+00 & 1.58E+00 & 1.58E+00 & 1.15E+00 & 1.15E+00 & 2.27E+01 & 2.23E+01 & 9.21E-01 & 9.22E-01 \\
           & Min  & 2.47E+00 & 2.43E+00 & 1.57E+00 & 1.56E+00 & 1.15E+00 & 1.14E+00 & 2.23E+01 & 2.15E+01 & 9.20E-01 & 9.18E-01 \\
           & Max  & 2.52E+00 & 2.54E+00 & 1.59E+00 & 1.59E+00 & 1.15E+00 & 1.16E+00 & 2.29E+01 & 2.28E+01 & 9.22E-01 & 9.24E-01 \\
           & Std  & 1.80E-02 & 5.18E-02 & 5.68E-03 & 1.64E-02 & 4.23E-03 & 1.05E-02 & 2.50E-01 & 5.43E-01 & 5.70E-04 & 2.29E-03 \\
           \hlineB{4}
CMAES-LSTM & Mean & 2.46E+00 & 2.46E+00 & 1.57E+00 & 1.57E+00 & 1.14E+00 & 1.14E+00 & 2.15E+01 & 2.18E+01 & 9.30E-01 & 9.28E-01 \\
           & Min  & 2.44E+00 & 2.40E+00 & 1.56E+00 & 1.55E+00 & 1.14E+00 & 1.13E+00 & 2.11E+01 & 2.13E+01 & 9.30E-01 & 9.25E-01 \\
           & Max  & 2.48E+00 & 2.51E+00 & 1.57E+00 & 1.58E+00 & 1.14E+00 & 1.15E+00 & 2.18E+01 & 2.22E+01 & 9.31E-01 & 9.31E-01 \\
           & Std  & 1.59E-02 & 5.49E-02 & 5.02E-03 & 1.74E-02 & 3.92E-03 & 9.00E-03 & 3.65E-01 & 4.19E-01 & 6.42E-04 & 2.63E-03\\
\hlineB{4}
\end{tabular}
}
\label{table:stat_onehour}
\end{table}
\section{Conclusions}
\label{sec:conclusion}
Wind speed forecasting plays an essential role in the wind energy industry. In this paper, we introduce a hybrid evolutionary deep learning approach (CMAES-LSTM) to acquire highly accurate and more stationary wind speed forecasting results. For tuning the LSTM network hyper-parameters, we propose two different techniques, a grid search and a well-known evolutionary method (CMA-ES). 
We evaluated the effectiveness of our approach we use data from the Lillgrund offshore wind farm and we use 10-minute and 1-hour time horizons.

Our experiments show that our approach outperforms others using all five performance indices, and that the performance difference is statistically significant. 

In the future, we are going to develop the proposed hybrid model by applying a decomposition approach to divide the time series wind speed data to some sub-groups which have more interrelated features. Then each sub-group is absorbed by one independent hybrid method to learn the nonlinear model of wind speed. Another future plan can be further investigations to compare the efficiency of different hybrid evolutionary algorithms and deep learning model based on the nonlinear combined mechanism. 
\section*{Acknowledgements}
 The authors would like to thank the Vattenfall project for the access to the SCADA data of the Lillgrund wind farm which is used in this study. Furthermore, This study is supported with supercomputing resources provided by the Phoenix HPC service at the University of Adelaide.


\bibliographystyle{unsrt}  
\bibliography{sample-bibliography}
\end{document}